\documentclass[10pt,twocolumn,letterpaper]{article}

\usepackage[pagenumbers]{cvpr} 
\usepackage{capt-of,etoolbox}
\usepackage{blindtext}
\usepackage{graphicx}
\usepackage{caption}
\usepackage[pagebackref,breaklinks,colorlinks,allcolors=cvprblue]{hyperref}
\usepackage[font=normalfont,labelfont=bf]{caption}
\usepackage{lipsum}
\usepackage{xcolor}
\usepackage{tikz}
\usetikzlibrary{shapes.geometric}
\usepackage[dvipsnames]{xcolor}
\usepackage{pifont} 
\usepackage{multirow}
\usepackage{tabularx,booktabs}
\usepackage[T1]{fontenc}  
\usepackage[noabbrev,capitalize]{cleveref}
\crefname{table}{Table}{Tables}
\usepackage[titletoc]{appendix}

\newcommand\score[1]{%
  \tikzstyle{scorestars}=[star, star points=5, star point ratio=2.25, draw, inner sep=1.3pt, anchor=outer point 3]%
  \begin{tikzpicture}[baseline]
    \draw (0, 0) node[name=star, scorestars, fill=cerulean]  {};
  \end{tikzpicture}%
}

\definecolor{cvprblue}{rgb}{0.21,0.49,0.74}

\def\paperID{6990} 
\def\confName{CVPR}
\def\confYear{2025}

\newcommand{\algo}{\textsc{SamIC}\xspace}
\newcommand{\cat}{\textsc{SamBox}\xspace}

\title{\textsc{SamIC}: Segment Anything with In-Context Spatial Prompt Engineering}

\author{Savinay Nagendra $^{1,2}$ \quad Kashif Rashid $^{2}$  \quad Chaopeng Shen$^{3}$ \quad Daniel Kifer$^{1}$ \vspace{0.3em} \\
{\normalsize $^1$Department of Computer Science and Engineering} \quad
{\normalsize $^3$Department of Civil and Environmental Engineering} \quad \\
{\normalsize The Pennsylvania State University, University Park, Pennsylvania} \\ {\normalsize $^2$Schlumberger-Doll Research, Cambridge, Massachusetts}\\
{\tt\small\centering sxn265@psu.edu \quad krashid@slb.com \quad cshen@engr.psu.edu \quad dkifer@cse.psu.edu \vspace{0.3em}}}

\begin{document}

\twocolumn[{%
\renewcommand\twocolumn[1][]{#1}%
\maketitle
\begin{center}
    \centering
    \captionsetup{type=figure}
    \includegraphics[width=0.96\textwidth]{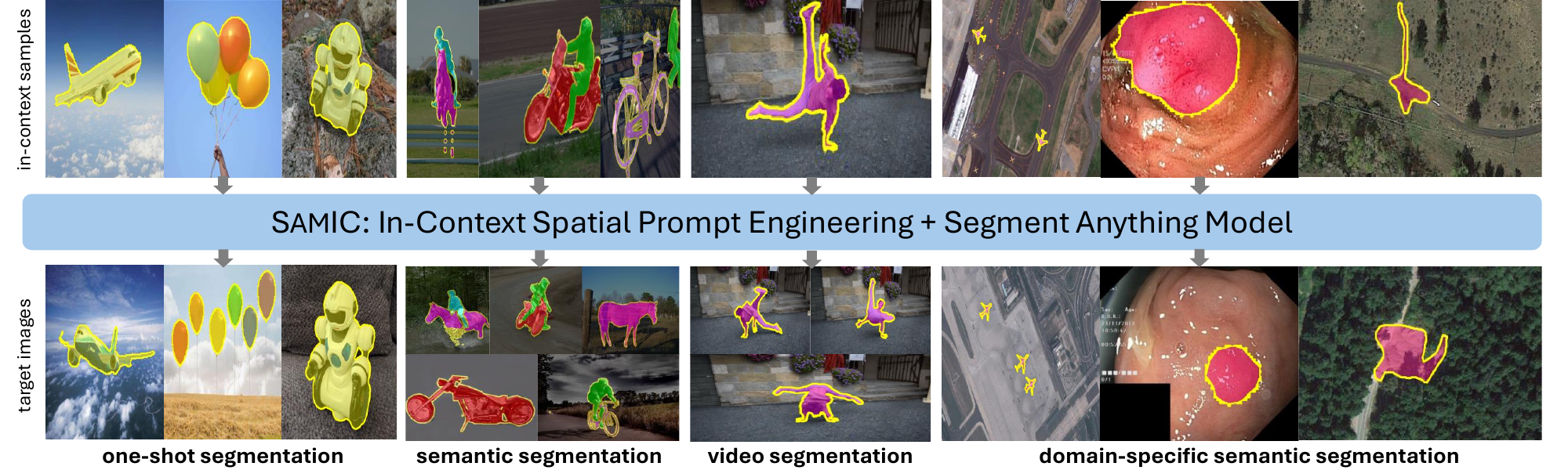}
    \captionof{figure}{\textbf{Qualitative results of \algo across diverse downstream segmentation tasks.} \algo has two components, the in-context spatial prompt engineering module that predicts task-specific spatial prompts for target images (bottom row) by learning from a few in-context samples (top row), and the Segment Anything Model (SAM) that takes the spatial prompts as input to produce valid masks. \algo unifies a diverse set of downstream segmentation tasks including one-shot segmentation, semantic segmentation, video object segmentation and domain-specific semantic segmentation.}
    \label{fig:overview_results}
\end{center}%
}]

\maketitle

\begin{abstract}
    Few-shot segmentation is the problem of learning to identify specific types of objects (e.g., airplanes) in 
    images from a small set of labeled reference images. The current state of the art is driven by resource-intensive construction of models for every new domain-specific application. Such models must be trained on enormous labeled datasets of unrelated objects (e.g., cars, trains, animals) so that their ``knowledge'' can be transferred to new types of objects. In this paper, we show how to leverage existing vision foundation models (VFMs) to reduce the incremental cost of creating few-shot segmentation models for new domains. Specifically, we introduce \algo, a small network that learns how to prompt VFMs in order to segment new types of objects in domain-specific applications. \algo enables any task to be approached as a few-shot learning problem. At 2.6 million parameters, it is 94\% smaller than the leading models (e.g., having ResNet 101 backbone with 45+ million parameters). Even using 1/5th of the training data  provided by one-shot benchmarks, \algo is competitive with, or sets the state of the art, on a variety of few-shot and semantic segmentation datasets including COCO-$20^i$, Pascal-$5^i$, PerSeg, FSS-1000, and NWPU VHR-10.

\end{abstract}

\section{Introduction}\label{sec:intro}


Semantic segmentation is the task of identifying the pixels in images and videos belonging to specific objects (like polyps in medical datasets, airplanes in aerial images, etc.). It is not possible to create comprehensive training datasets, with pixel-level labeling, for all possible objects in existence. Hence, there is considerable interest in models that learn from small datasets, or even perform few-shot learning, where a model trained on one dataset is taught to recognize new types of objects from as few as 1-2 labeled reference images.

\begin{figure}[htpb]
    \centering
    \includegraphics[width=\linewidth]{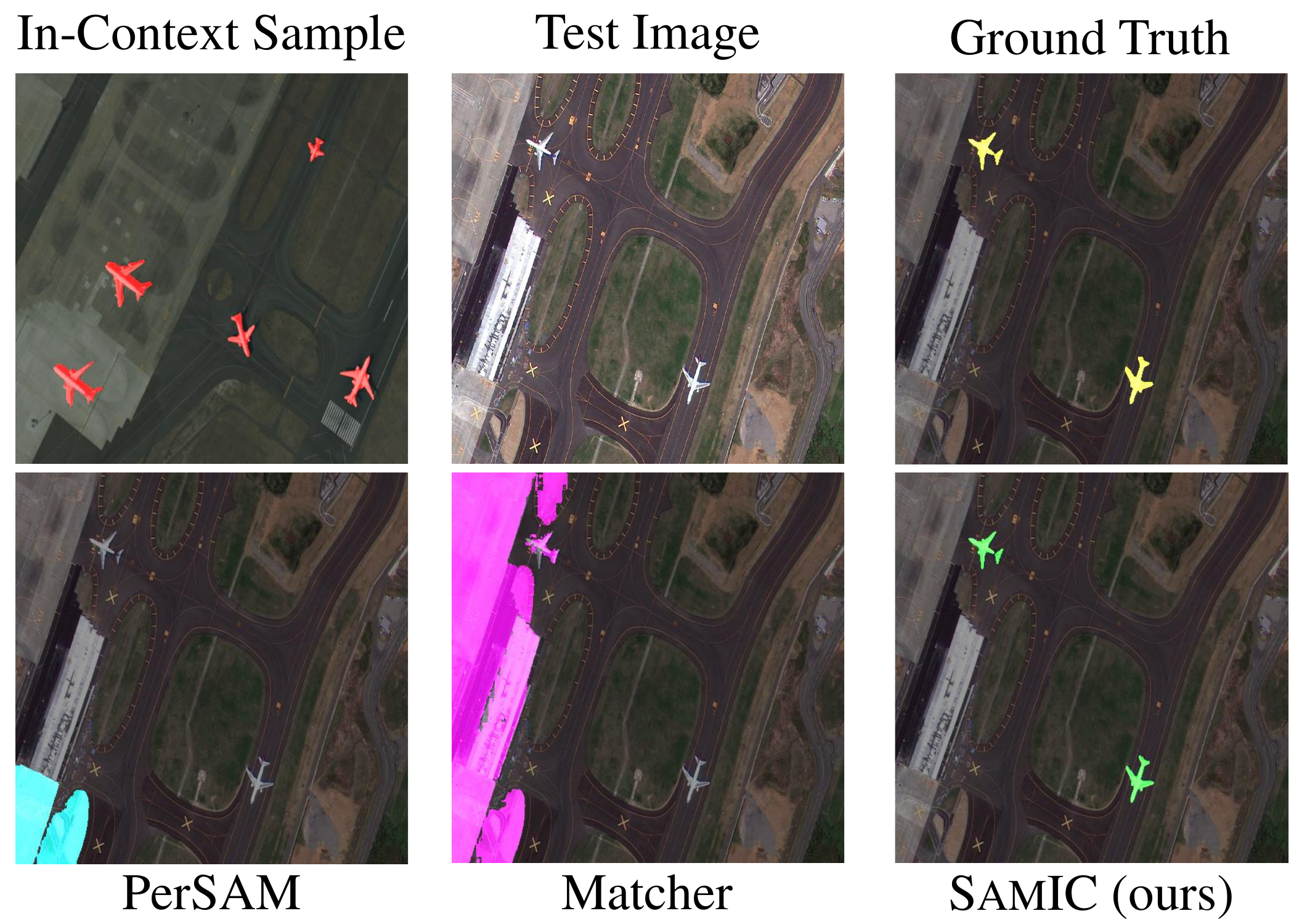}
    \caption{Comparison of top prompt engineering methods (PerSAM \cite{zhang2023personalize}, Matcher \cite{liu2023matcher}) for 1-shot airplane segmentation on a sample from NWPU VHR-10 \cite{su2019object}. The reference image contains masks of airplanes. The test image has similar airplanes, but the orientation and background differ.
    }
    \label{fig:intro-comp}
\end{figure}

Many state-of-the-art few-shot benchmarks are dominated by relatively large models \cite{chen2024visual, hong2021cost, zhang2022feature} (e.g., $45+$ million parameters) that are trained on massive datasets \cite{lin2014microsoft, shaban2017one, li2020fss}. As a result, adding support for new domains is resource-intensive in terms of hardware and energy expenditures (e.g., one-shot segmentation models for animals need to be retrained to handle vehicles). On the other hand, the existence of Vision Foundation Models (VFMs) like the Segment Anything Model (SAM) \cite{kirillov2023segment} opens up the possibility that new domains can be supported by in-context learning (ICL) -- specifically, prompt engineering. That is, much smaller models, trained on much smaller datasets, learn how to interact with a VFM to properly segment new classes of objects. In this spirit, we propose, a lightweight model named {\textbf{\algo}}\footnote{Source code will be released upon acceptance.} that is competitive with, and often outperforms, much larger models (and existing prompt engineering methods \cite{zhang2023personalize, liu2023matcher}) on few-shot and small data benchmarks \cite{shaban2017one, li2020fss, jha2020kvasir, pont20172017, nagendra2022constructing, su2019object}.

VFMs \cite{oquab2023dinov2, kirillov2023segment, ilharco2021openclip,minderer2022simple} now serve as backbones for multiple vision tasks, such as classification \cite{woo2023convnext}, object detection \cite{terven2023comprehensive, zhu2020deformable, minderer2022simple}, and segmentation \cite{kirillov2023segment, oquab2023dinov2, cheng2022masked}. Given good \emph{manual} prompts, they excel at many applications such as stable diffusion \cite{rombach2022high}, visual query answering \cite{antol2015vqa}, and semantic/instance/panoptic segmentation \cite{csurka2022semantic, hafiz2020survey, kirillov2019panoptic}. In particular, SAM \cite{kirillov2023segment} is a VFM that can segment objects when manually provided with spatial prompts  like pixel coordinates, bounding boxes, or raw masks.

Thus, instead of designing new custom models for computer vision tasks like semantic segmentation, one can turn it into a prompt engineering problem -- training a model to give good prompts to SAM. 
Previous attempts at using ICL for segmentation \cite{wang2023images, wang2023seggpt, liu2023matcher, tang2023emergent, zhang2023personalize} have had mixed success. The most effective thus far are training-free spatial prompt engineering methods -- PerSAM \cite{zhang2023personalize} and Matcher \cite{liu2023matcher} that leverage frozen VFM encoders \cite{oquab2023dinov2, kirillov2023segment} to extract embeddings. They create prompts in a new image by finding regions whose embeddings are similar to the labeled objects
in the reference image. However, embeddings are complex and opaque objects that can capture information that may be irrelevant to a specific task. For these reasons, the training-free methods tend to be brittle -- they can be distracted by changes in color, spatial orientation, and presence of other objects. An example is shown in \cref{fig:intro-comp}.

To get around this brittleness, methods need to be trained to understand what embeddings mean so that they can be used more effectively to learn about object boundaries, distinguishing instances, or grouping semantic regions. Thus \algo requires a small dataset of labeled objects to build ``knowledge'' it can transfer to new types of objects. Afterwards it can analyze one reference image containing new types of objects and design prompts for future images. 
%
%
%
Inspired by the success of the lightweight Hypercorrelation Squeeze Network (HSNet) \cite{min2021hypercorrelation} in few-shot learning, we adopt HSNet as the underlying architecture for \algo and we also created a custom annotation tool called \textbf{\cat}. 

\algo and \cat operate as follows. To create an object mask for a reference image, a user would employ \cat to add prompts to the reference image. \cat displays the segmentation produced by SAM and allows the user to correct the prompts until the object is properly segmented. This is much faster than traditional annotation tools.
\algo then places 2D Gaussians around these point prompts to create saliency-like heat maps \cite{jia2020eml} as a data pre-processing step. When new testing images start coming in, \algo takes that reference image, its  heat map, and a new testing image, and then predicts the heat map for the testing image. Peaks in the predicted heatmaps are identified using a peak detection algorithm that we devised. These peaks become the prompts to the testing image, and are sent to SAM for segmentation.

\noindent Our main contributions are summarized as follows:
\begin{itemize}
    \item We present \algo, a simple and highly efficient spatial prompt engineering framework that learns how to generate task-specific spatial point prompts for SAM.  \algo uses a novel in-context saliency prediction architecture for predicting  heat maps from which spatial point prompts are extracted using a peak detection algorithm.
    \item \algo is paired with a user-friendly annotation tool we call \cat to facilitate the rapid collection of spatial point prompts for in-context learning. When we were preparing experiments for this paper, we found that using \cat we could annotate images 6$\times$ faster than with traditional tools like LabelMe \cite{kentaro}.
    \item \algo uses an extremely small architecture (approximately 1/20$^\text{th}$ the size of other state-of-the-art architectures for 1-shot segmentation), making it very cost-efficient to train and deploy in new applications. 
    \item \algo outperforms prior prompt-engineering methods and achieves state-of-the-art performance on a range of downstream tasks. To emphasize its data efficiency, \algo is only given 20\% of the training data provided by the one-shot benchmarks, but is compared against methods that use all of the training data. For instance, \algo achieves state-of-the-art performance on \footnote{Note that, for the Pascal=$5^i$ dataset, we do not consider SegGPT \cite{wang2023seggpt}, because it trains on the testing data for Pascal-VOC dataset.}{Pascal-$5^i$} \cite{shaban2017one}, and PerSeg \cite{zhang2023personalize} datasets with 1-shot mIoUs of 80.4\%, and 97.5\%, surpassing the previous state-of-the-art FP-Trans (ViT-B) \cite{zhang2022feature} and Matcher \cite{zhang2023personalize} by +2.1\% and +10.9\% respectively. Additionally, \algo significantly surpasses Matcher \cite{liu2023matcher}, the previous state-of-the-art spatial prompt engineering method on domain-specific, real-world applications including gains of +35.3\% on TSMU \cite{nagendra2022constructing}, +5.8\% on Kvasir-SEG \cite{jha2020kvasir} and +18.5\% on NWPU VHR-10 \cite{su2019object} datasets. On FSS-1000, \algo achieves 1-shot mIoU of 89\%, while the state-of-the-art method VAT (ResNet-101) \cite{zhang2022feature} with a much more complex model trained on 100\% of the data achieves 90.3\%.
    %

\end{itemize}

\section{Related Work} \label{sec:related_work}
\textbf{Vision Foundation Models (VFMs).} In recent years, large-scale vision pre-training has driven significant advancements in VFMs, forming the basis for many state-of-the-art techniques \cite{bao2021beit, caron2021emerging, chen2020simple, he2022masked, he2020momentum, kirillov2023segment, radford2021learning, jia2021scaling,li2021align, radford2021learning,rombach2022high,yu2022coca,nagendra2022constructing, funk2018learning, pei2021utilizing, nagendra2020cloud, nagendra2022threshnet, nagendra2023estimating}
Vision-based models focus on tasks like distinguishing image or patch-level entities from different perspectives \cite{caron2021emerging, chen2020simple, he2020momentum, nagendra2017comparison, nagendra2024patchrefinenet, liu2021new, nagendra2020efficient, zhu2022rapid, nagendra2024emotion, nagendra2024thermal} or reconstructing masked portions of images \cite{bao2021beit, he2020momentum}. For example, DINOv2 \cite{oquab2023dinov2}, a self-supervised VFM, surpasses OpenCLIP \cite{ilharco2021openclip} in general-purpose feature extraction and exhibits strong visual feature matching. 
Of particular note, SAM \cite{kirillov2023segment} is a promptable VFM, pre-trained on the SA-1B dataset comprising 1B masks and 11M images, and uses precise spatial prompts to guide the model’s learning. 

\noindent\textbf{In-Context Segmentation.} Recent research \cite{bai2024sequential, bar2022visual, liu2023matcher, wang2023images, wang2023seggpt} has explored the use of in-context learning for segmentation tasks. Approaches like Painter \cite{wang2023images} and SegGPT \cite{wang2023seggpt} implement in-context segmentation through image in-painting, building on the Masked Image Modeling (MIM) framework \cite{he2022masked}. These models concatenate images and predictions into a 2×2 mosaic, predicting by reconstructing the masked regions. However, this process relies on a vision backbone that functions as both an image encoder and mask decoder, leading to high computational demands. Additionally, these methods face challenges in effectively utilizing pre-trained models due to input inconsistencies, resulting in slower convergence. PerSAM \cite{zhang2023personalize} and Matcher \cite{liu2023matcher} take a different approach by leveraging SAM \cite{kirillov2023segment} for in-context segmentation via prompting. They create one-shot, cross-image correspondences between examples and targets by computing cosine similarity matrices from generic VFM embeddings. However, these methods struggle to adapt to a wide range of downstream tasks due to the limitations of their prompt generation strategies. In contrast, \algo addresses these shortcomings by learning visual correspondences between in-context samples and target images, allowing for the prediction of task-specific spatial prompts.

\section{Segmentation with \algo}\label{sec:approach}
%
The VFM SAM has a class-agnostic design \cite{zhang2023personalize} and therefore requires prompts in order to disambiguate what are the objects of interest and which image patches belong to the same (or different) objects. 
This presents an opportunity for models that learn \emph{how} to prompt a VFM and is the motivation for \algo, a simple and efficient automated prompt engineering mechanism. 

An RGB image or video frame with height $H$ and width $W$ can be represented as a tensor $\mathbf{I}\in \mathbb{R}^{3 \times H \times W}$. We let $\mathbf{I}(x,y)$ represent the RGB triple at location $(x,y)$ in the image. 
The superscript $t$ indicates that an image $\mathbf{I}^t$ is a target/testing image. A spatial prompt $\mathbf{P}$ is a set of points. An in-context sample is a set of $K$ images and their associated spatial prompts: $\{(\mathbf{I}^c_1, \mathbf{P}_1), \dots, (\mathbf{I}^c_K, \mathbf{P}_K)\}$ where the spatial prompt $\mathbf{P}_i=\{(x_1, y_1), \dots, (x_{N_i}, y_{N_i})\}$ contains $N_i$ points. For one-shot learning, $K=1$.  In a typical use-case, the in-context sample contains examples of a new type of object. 
The goal is to use these in-context samples to figure out how to automatically add spatial prompts to future images $\mathbf{I}^t_1, \mathbf{I}^t_2, \dots$ that contain such objects so that they can be segmented without any further human annotation.


\subsection{Collecting unambiguous user prompts}\label{sec:gt-generation}
\begin{figure}
\centering
    \includegraphics[width=\linewidth]{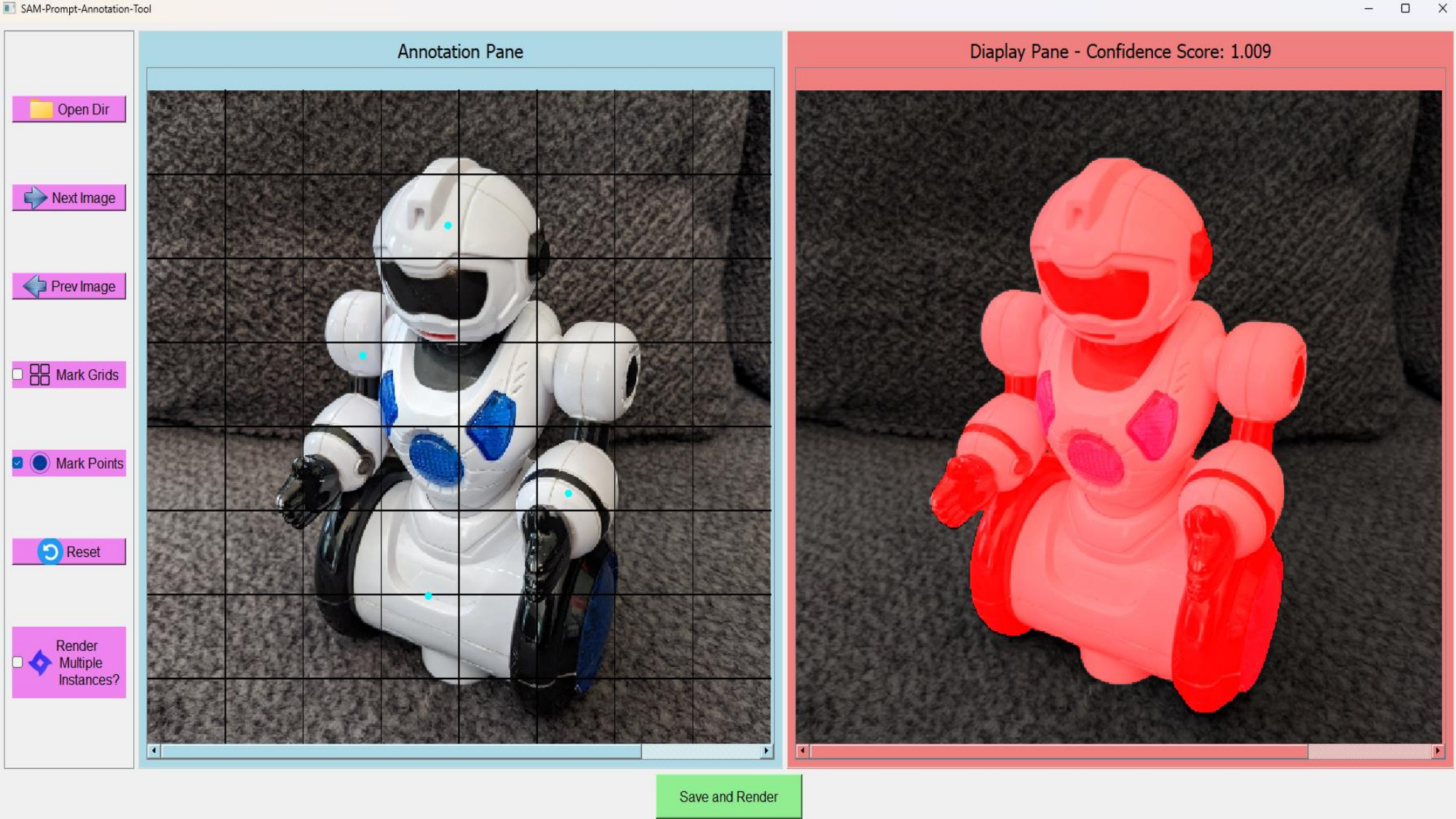}
    \caption{The  \cat annotation tool for rapid collection of unambiguous user prompts. With four user-provided spatial point prompts, \cat outputs a mask with saturated confidence score of 1.009}
    \label{fig:tool}
\end{figure}

 To stream-line the annotation of in-context samples with good spatial prompts, we built an interactive annotation tool named \cat (a snapshot of the tool is shown in \cref{fig:tool} with cyan colored point prompts). The purpose of \cat is to guide the user into generating high quality, unambiguous prompts that will later provide a strong learning signal for \algo. \cat interfaces with SAM and addresses some of its front-end limitations, especially with the segmentation of disjoint object instances in an image. For example, if one annotates two airplanes in an image using SAM's front-end, SAM would return a segmentation that also includes the runway linking the airplanes together (\algo will also need to work around this limitation when predicting prompts).
%
%
First, \cat takes all of the images the user wants to annotate and runs them through SAM's image encoder, saving the results to disk (this is a one-time pre-processing cost and takes $\approx 20$ seconds per image on a Nvidia RTX 2080 Ti GPU). Then the user annotates each image sequentially via \cat. If the user only wishes to annotate one object in an image, they would provide some point prompts or boxes and press ``Save and Render''. \cat sends these prompts to SAM, which responds with a mask and an ambiguity-aware confidence score.\footnote{For example, if the prompt is on a shirt, SAM does not know if the intention is to segment the shirt or the entire person wearing it, so the confidence score would be low.} This information is presented to the user, who can then provide additional prompts, if necessary, to correct the segmentation.

The ambiguity score is critical to the success of \algo in one-shot segmentation. Even if SAM correctly segments an object, the user could keep adding point prompts until the confidence is nearly 1, as that represents a good, unambiguous prompt that \algo can learn from.

If the user wishes to segment multiple instances of an object in an image, \cat would guide the user into providing prompts for each object. Prompts for each object are sent separately to SAM, to get multiple masks, which are then unioned by \cat.

\begin{figure}
    \centering
    \includegraphics[width=\linewidth]{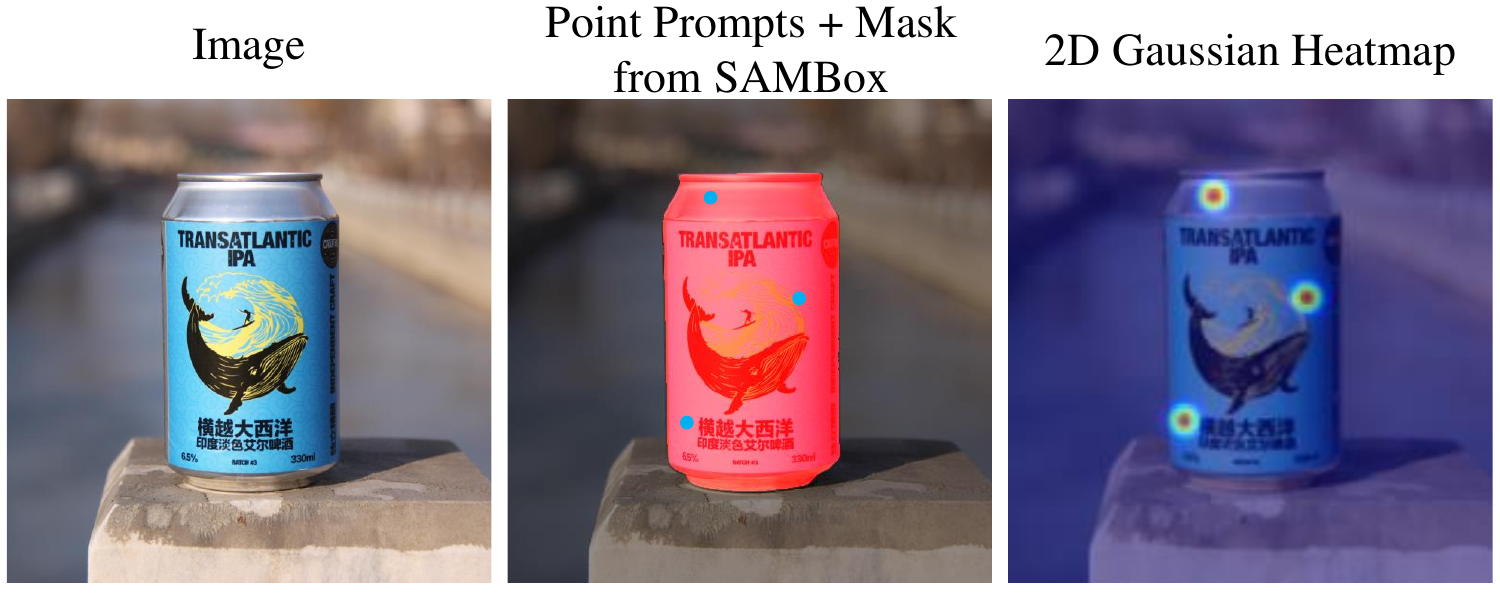}
    \caption{ Saliency-like 2D Gaussian heatmap generated from user-provided prompt with \cat.}
    \label{fig:saliency_heatmap}
\end{figure}

Collected point prompts are stored in a JSON file. Additionally, pixel-level masks are also stored, enabling the tool to function as a stand-alone rapid annotation platform for segmentation tasks. The tool supports instance and semantic segmentation tasks, making it highly scalable and adaptable for a wide range of applications. Using a standard RTX 2080 Ti GPU, a typical image can be annotated in less than 60 seconds, which, for us, was $\approx 5\times$ faster than manual pixel-wise annotation using a typical tool such as LabelMe \cite{kentaro}. For instance-segmentation or domain-specific tasks (where multiple objects have to be annotated), the tool accelerated our work  by up to $\approx 15\times$. 

\begin{figure*}[ht]
    \centering
    \includegraphics[width=0.8\linewidth]{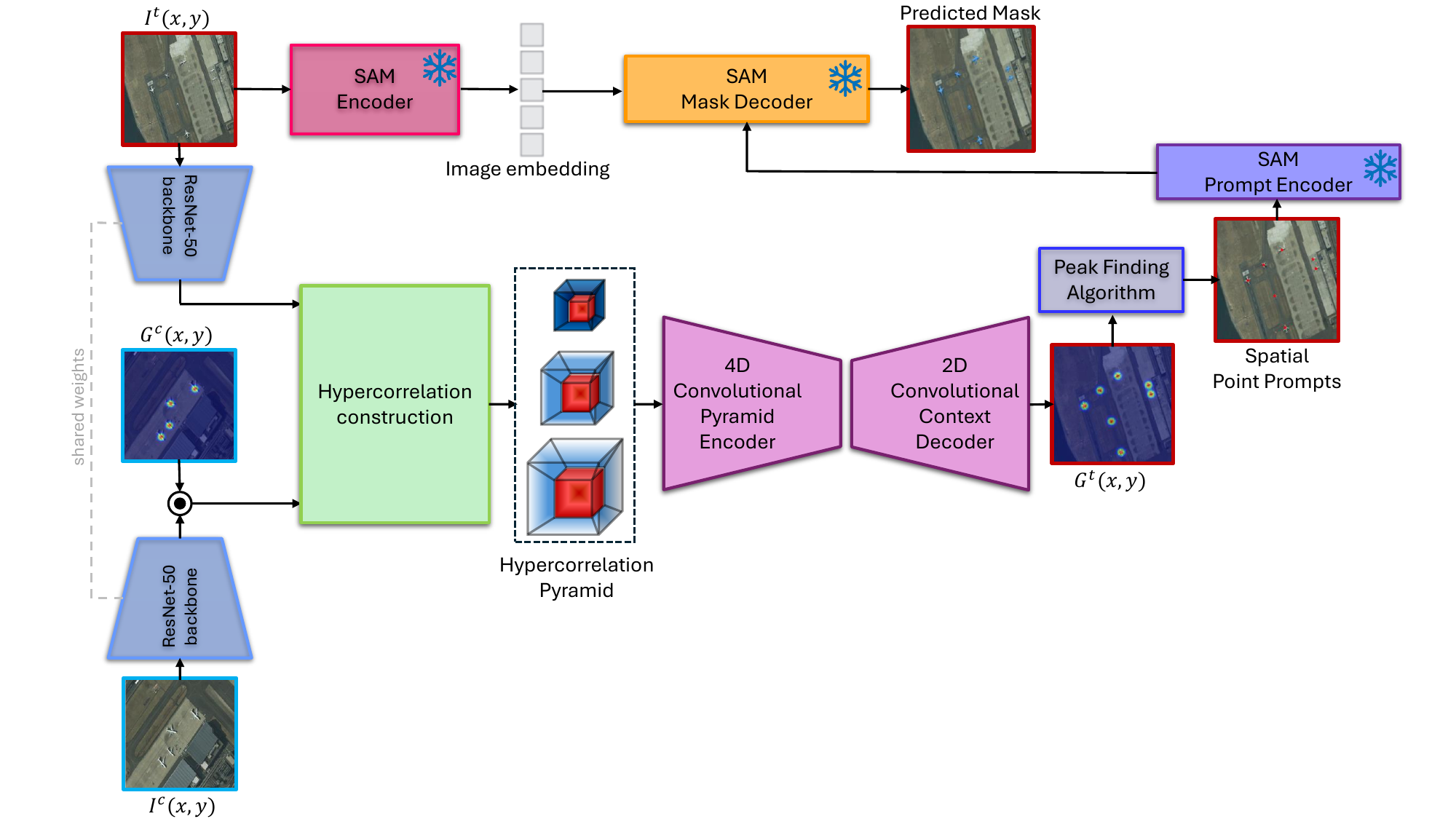}
    \caption{\textbf{Overview of the \algo architecture. }HSNet \cite{min2021hypercorrelation} is used as our in-context visual saliency prediction architecture to predict a saliency-like heat map with 2D Gaussians, representing location priors for task-specific spatial point prompts. A peak finding algorithm is used to extract a sequence of point prompts from the predicted heat map that are provided to SAM to generate segmentation masks.}
    \label{fig:samic_overview}
\end{figure*}


\subsection{Point Prompts to Saliency Heat Maps}\label{sec:heat}
Spatial point prompts are like indicators of the saliency of components of an object of interest.
Thus, drawing inspiration from visual saliency prediction \cite{jia2020eml}, we represent the spatial point prompts $\{(x_1, y_1), (x_2, y_2), ..., (x_N, y_N)\}$ for an in-context image $\mathbf{I}^c$ as a saliency-like heat map $\mathbf{G}$ with the same height $H$ and width $W$ as $\mathbf{I}^c$. The heat map consists of Gaussians centered around each prompt. The intensity of the heatmap at point $(x,y)$ is:
\begin{align*}
    \mathbf{G}(x,y) =  \sum_{k=1}^{N}\exp\left(-\frac{\left(\frac{x-x_k}{W}\right)^2 + \left(\frac{y-y_k}{H}\right)^2}{2\sigma^2}\right),
\end{align*}
where $\sigma$ controls the diffuseness of the Gaussians (we use $\sigma=0.02$ throughout our experiments). 
The heatmap is then normalized (divided by the max value) to ensure every entry is $\in [0,1]$.
An example heatmap in shown in \cref{fig:saliency_heatmap}.


\subsection{\algo Architecture}\label{sec:arch}

The input to \algo is an in-context image with a heatmap along with a new testing image, and the output is a heat map for the testing image. When there are $K>1$ in-context images, they are fed to \algo one at a time with the testing image to get $K$ heatmaps, which are then averaged. Then a peak-finding algorithm converts the heatmap into prompts (an overview of the process is shown in Figure \ref{fig:samic_overview}). In principle, any architecture could be used with \algo, but we selected the Hypercorrelation Squeeze Network (HSNet) \cite{min2021hypercorrelation} (rather than, say, a transformer-based model \cite{lou2022transalnet}) because it is lightweight and has shown success in other few-shot learning applications \cite{li2020fss}. For completeness, we briefly review the HSNet architecture.


\noindent\textbf{Hypercorrelation Construction.} 
Inspired by previous semantic matching approaches \cite{liu2020semantic, min2019hyperpixel}, HSNet \cite{min2021hypercorrelation} leverages a comprehensive set of features from intermediate layers of a convolutional neural network to capture both semantic and geometric patterns of similarity between in-context and target images. In all experiments, ResNet-50 \cite{he2016deep} is selected as the backbone network, which produces a sequence of $L$ pairs of feature maps, denoted as ${F^c_l, F^t_l}_{l=1}^{L}$, for the reference and target images $\mathbf{I}^c$ and $\mathbf{I}^t$, respectively. Here, $F^c_l$ and $F^t_l$ represent the feature maps obtained from the $l^{th}$ layer of the network. To focus on relevant activations, each feature map for the in-context image is masked with a saliency-like heatmap: $\hat{F}_l^c = F^c_l \odot \zeta(G^c)$, where $\odot$ denotes the Hadamard product and $\zeta(.)$ is a function that bilinearly interpolates the input tensor to match the spatial dimensions of the feature map. 4D correlation tensors between the two sets of feature maps are computed using cosine similarity: $\hat{C}_l(x^c,x^t) = ReLU\left(\frac{F^c_l(x^c) \cdot F^t_l(x^t)}{\left \lVert F^c_l(x^c)\right \rVert \left \lVert F^t_l(x^t)\right \rVert}\right)$, where $x^c$ and $x^t$ represent the 2D spatial positions in $F^c_l$ and $F^t_l$. These 4D correlation tensors are concatenated along the channel dimension to create a hypercorrelation tensor for each layer $l$, denoted as $C_l \in \mathbb{R}^{|L_l|\times H_l \times W_l\times H_l \times W_l}$. This process is repeated across different layers to obtain hypercorrelations at various spatial resolutions and depths, which are stacked together to form the hypercorrelation pyramid $C$. 

\noindent\textbf{4D-Convolutional Pyramid Encoder. }
The hypercorrelation pyramid $C$ is compressed into a more compact feature map through the use of two types of building blocks: a squeezing block and a mixing block. Each block comprises three stages: multi-channel 4D convolution \cite{choy20194d}, group normalization \cite{wu2018group}, and ReLU \cite{agarap2018deep} activation. In the squeezing block, large strides are applied to reduce the last two spatial dimensions of $C$ while preserving the first two dimensions. The outputs from adjacent layers in the pyramid are combined using element-wise addition after upsampling the spatial dimensions of the upper layer. The mixing block then processes this combined output with 4D convolutions, facilitating the flow of relevant information from higher to lower layers in a top-down manner. Lastly, average pooling is applied to produce the condensed representation $Z$ of the hypercorrelation pyramid $C$.

\noindent\textbf{2D-Convolutional Context Decoder}
The decoder network consists of a series of 2D convolutions, ReLU, and upsampling layers followed by softmax \cite{bridle1989training} function. The network takes
the context representation $Z$ and predicts $G^t \in [0,1]^{1\times H \times W}$, which is the predicted saliency-like heat map for the target image $I^t$. 

\subsection{Post-Processing: From Heatmaps to Prompts}\label{sec:pp}

We devise a peak detection algorithm to extract spatial point prompts $P^t$ from a predicted saliency-like heatmap $\widehat{G}^t$. 
This algorithm works as follows: First, $\widehat{G}^t$ is binarized with threshold $\tau=0.5$ to get a binary map $B^t \in \{0,1\}$. Second, contours are extracted from $B^t$ using a contour detection algorithm \cite{suzuki1985topological} to identify connected components $R = \{R_1, R_2, ..., R_N\}$ in $B^t$. 
For each connected component $R_i$, we calculate spatial moments \cite{hu1962visual} $M_{x,i}$, $M_{y,i}$, and the area $A_i$, where $M_{x,i} = \sum_{(x,y)\in R_i}x  B_t(x,y)$, $M_{y,i} = \sum_{(x,y)\in R_i}y  B_t(x,y)$ and $A_i = \sum_{(x,y)\in R_i}B_t(x,y)$. We then calculate the centroid $(x_i, y_i)$ of each connected component $R_i$ as $x_i = \frac{M_{x,i}}{A_i}$, $y_i = \frac{M_{y,i}}{A_i}$, which represents the peak of $G_i$. Finally, the extracted peaks $P^t = \{(x_1,y_1), (x_2,y_2), ..., (x_N,y_N)\}$ become the spatial point prompts provided to SAM.

\subsection{Loss Functions For Training}\label{sec:loss-function}
When training, let us denote the predicted and ground truth saliency-like heat maps as $\widehat{G}$ and $G$, respectively. 
Our loss function, adopted from standard visual saliency prediction methods \cite{lou2022transalnet, jia2020eml}, has three components:
\\
\noindent \textbf{Kullback-Leibler Divergence (KLD) }  \cite{bylinskii2018different} measures the similarity between the two heat maps. 
\begin{equation}
    Loss_{KLD}(G, \widehat{G}) = \sum_i G_i log\left(\epsilon + \frac{G_i}{\widehat{G}_i + \epsilon}\right)~,
\end{equation} where $i$ denotes the location of pixels in the heat map, and $\epsilon = 1e^{-6}$ is a  term that prevents division by 0. \\
\noindent \textbf{One Minus Pearson’s Correlation Coefficient (CC) } \cite{cohen2009pearson} is a statistical method that measures the linear correlation between two random variables. 
\begin{equation}
 Loss_{CC} = 1 -   CC(\widehat{G},G) = 1 - \frac{\sigma(\widehat{G},G)}{\sigma(\widehat{G}) \times \sigma(G)}~,
\end{equation} where $\sigma(\widehat{G},G)$ denotes the covariance of $\widehat{G}$ and $G$ while $\sigma(\widehat{G})$ and $\sigma(G)$ are the standard deviations of $\widehat{G}$ and $G$, respectively. This loss function takes values in $[0, 2]$.\\
\noindent\textbf{Normalized Scanpath Saliency(NSS)} \cite{peters2005components} is used to measure the average normalized saliency between two binary fixation maps, which are defined as follows. Let $F$ denote the binarized version of the ground truth heat map $G$ with threshold $0.5$. That is, $F(x,y) = \mathbf{1}_{G(x,y)\geq 0.5}$. $F$ is known as the fixation map.
\begin{equation}
    Loss_{NSS}(\widehat{G}, F) = \frac{1}{N}\sum_i(\Bar{R}_i - \Bar{\widehat{G}}_i) \times F_i~,
\end{equation} where $N = \sum_i F_i$ and $\Bar{\widehat{G}}_i = \frac{\widehat{G} - \mu(\widehat{G})}{\sigma(\widehat{G})}$ and $\Bar{R} = \frac{G - \mu(G)}{\sigma(G)}$. \\
\noindent\textbf{The final loss function} is a summation of all the components: $Loss = Loss_{KLD} + Loss_{CC} + Loss_{NSS}$.

\section{Experiments}\label{sec:experiments}
VFMs and large-scale pre-trained models have created problems for the fair evaluation of  computer vision algorithms on standard benchmarks, as these models are often trained on the entire benchmarks (including the testing sets). This can skew the results for methods built on top of them. This caveat is true for competing methods
and also for \algo because it is not known exactly what the VFM SAM is trained on. However, based on dataset recency, we suspect that PerSeg \cite{zhang2023personalize} and TSMU \cite{nagendra2022constructing} are least likely to have been used by SAM or large-scale pre-training models. In addition, NWPU VHR-10  \cite{su2019object} and Kvasir-SEG  \cite{jha2020kvasir} are not cited by SAM \cite{kirillov2023segment} and so probably are not used by other pre-training models either. Thus,  these 4 datasets may offer the fairest comparisons. We use a $\textcolor{red}{\dagger}$ to note when a model has a known potentially unfair advantage.


\subsection{Datasets}
Our experiments span nine benchmark \footnote{For additional details about the processing and construction of the datasets, please see supplementary material.}{datasets} across four diverse tasks: few-shot segmentation, aerial segmentation, medical segmentation, and video object segmentation. Aerial and medical segmentation are classified as domain-specific segmentation tasks. For all datasets, \algo is only given \underline{$20\%$ of the training data} from each class. \\
\textbf{Few-shot Segmentation} is evaluated using the COCO-$20^i$ \cite{nguyen2019feature}, Pascal-$5^i$ \cite{shaban2017one}, FSS-1000 \cite{li2020fss}, and PerSeg \cite{zhang2023personalize} datasets. COCO-$20^i$ divides the 80 MSCOCO categories into four cross-validation folds, while Pascal-$5^i$ does the same for the 20 PASCAL VOC 2012 categories. FSS-1000 consists of 1,000 classes, each with 10 mask-annotated images. The dataset is split into 520 training classes, 240 for validation, and 240 for testing, supporting balanced and comprehensive evaluation across tasks. PerSeg is a personalized object detection dataset with 40 classes, which originates from Dreambooth \cite{ruiz2023dreambooth} and is annotated by \cite{zhang2023personalize}. We follow the standard evaluation scheme in \cite{min2021hypercorrelation}.

\noindent \textbf{Video Object Segmentation (VOS)}  targets  specific objects across video frames. We evaluate \algo on the test splits of DAVIS-16 \cite{perazzi2016benchmark} and DAVIS-17\cite{pont20172017}. For training, we apply optical flow \cite{horn1981determining} to each video to extract the most dynamic frames, sampling 10 frames per video ($\approx 20\%$ of the training data). Frames are structured to maintain causality, with frame $t-1$ used as reference for target frame $t$. Evaluation uses VOS metrics: $J$ and $F$ scores.

\noindent \textbf{Domain-specific Segmentation.} We evaluate \algo on domain-specific datasets that resemble real-world applications, rather than standard benchmarks or VFM pre-training datasets. We evaluate on two domain-specific tasks: (1) Aerial segmentation using the TSMU landslide \cite{nagendra2022constructing}, and NWPU VHR-10 \cite{su2019object} datasets, and (2) Medical segmentation using the Kvasir-SEG \cite{jha2020kvasir} dataset. 

\subsection{Implementation Details}
We employ a ResNet-50 backbone \cite{he2016deep} pre-trained on ImageNet \cite{russakovsky2015imagenet}. Features are extracted at the end of each bottleneck before the ReLU \cite{agarap2018deep} activation, yielding three pyramidal layers. The spatial size of reference and target images is set to $224 \times 224$ in our experiments, though it can be adjusted for custom datasets. The backbone network is frozen to prevent learning class-specific representations. \algo is trained with the Adam optimizer \cite{kingma2014adam}, using an initial learning rate of 1e-3 and a batch size of 4. With only $2.6M$ learnable parameters, training on any custom dataset with a single class converges within 300 epochs (approx. 20 minutes on an NVIDIA 2080 Ti GPU), and early stopping is applied if training loss does not improve over 10 epochs. 


\begin{table*}[htpb]
\centering
\resizebox{\linewidth}{!}{%
\begin{tabular}{l|c|ccccccc|cc}
\hline
\multicolumn{1}{c|}{\multirow{3}{*}{Methods}} &
  \multirow{3}{*}{\begin{tabular}[c]{@{}c@{}}\# learnable \\ parameters\end{tabular}} &
  \multicolumn{4}{c|}{few-shot segmentation} &
  \multicolumn{2}{c|}{aerial segmentation} &
  medical segmentation &
  \multicolumn{2}{c}{video object segmentation} \\ \cline{3-11} 
\multicolumn{1}{c|}{} &
   &
  \multicolumn{1}{c|}{COCO-$20^i$ \cite{nguyen2019feature}} &
  \multicolumn{1}{c|}{Pascal-$5^i$ \cite{shaban2017one}}  &
  \multicolumn{1}{c|}{FSS-1000 \cite{li2020fss}} &
  \multicolumn{1}{c|}{PerSeg \cite{ruiz2023dreambooth}} &
  \multicolumn{1}{c|}{TSMU \cite{nagendra2022constructing}} &
  \multicolumn{1}{c|}{NWPU VHR-10 \cite{su2019object}} &
  Kvasir-SEG \cite{kirillov2023segment} &
  \multicolumn{1}{c|}{DAVIS-16 \cite{pezoa2016foundations}} &
  DAVIS-17 \cite{pont20172017} \\ \cline{3-11} 
\multicolumn{1}{c|}{} &
   &
  \multicolumn{7}{c|}{mIoU (1-shot)} &
  \multicolumn{2}{c}{\textit{J \& F} (1-shot)} \\ \hline
PerSAM (ViT-H) \cite{zhang2023personalize} &
  0 &
  \multicolumn{1}{c|}{23.0} &
  \multicolumn{1}{c|}{29.8} &
  \multicolumn{1}{c|}{71.2} &
  \multicolumn{1}{c|}{86.1} &
  \multicolumn{1}{c|}{36.3} &
  \multicolumn{1}{c|}{23.1} &
  45.8 &
  \multicolumn{1}{c|}{\textcolor{gray}{\textcolor{gray}{n/a}}} &
  60.3 \\
Matcher (ViT-G) \cite{liu2023matcher} &
  0 &
  \multicolumn{1}{c|}{52.7} &
  \multicolumn{1}{c|}{76.9} &
  \multicolumn{1}{c|}{87.0} &
  \multicolumn{1}{c|}{86.6} &
  \multicolumn{1}{c|}{52.9} &
  \multicolumn{1}{c|}{51.6} &
  63.5 &
  \multicolumn{1}{c|}{86.1} &
  79.5 \\
\algo (ours) &
  2.6M &
  \multicolumn{1}{c|}{\textbf{53.1}} &
  \multicolumn{1}{c|}{\textbf{80.4}} &
  \multicolumn{1}{c|}{\textbf{89.0}} &
  \multicolumn{1}{c|}{\textbf{97.5}} &
  \multicolumn{1}{c|}{\textbf{88.2}} &
  \multicolumn{1}{c|}{\textbf{70.1}} &
  \textbf{69.3} &
  \multicolumn{1}{c|}{\textbf{87.2}} &
  \textbf{80.6} \\ \hline
\end{tabular}
}
\caption{Comparison of \algo with top spatial prompt engineering methods PerSAM \cite{zhang2023personalize} and Matcher \cite{liu2023matcher} on 9 benchmark datasets spanning four diverse tasks of few-shot segmentation, aerial segmentation, medical segmentation and video object segmentation. }
\label{tab:prompt_engieering}
\end{table*}

\subsection{Quantitative Results} \label{sec:results}
In our experiments, we evaluate methods from three categories: (i) \textit{specialist segmentation models} that are fully supervised and trained to convergence on specific datasets, (ii) \textit{in-context generalist segmentation models}, which are trained on major computer vision benchmarks and require in-context samples (reference image + reference mask) for segmentation, such as Painter \cite{wang2019panet} and SegGPT \cite{wang2023seggpt}, and (iii) \textit{in-context spatial prompt engineering models}, which generate spatial prompts and utilize SAM \cite{kirillov2023segment} as the segmenter. This third category includes (a) training-free methods such as PerSAM \cite{zhang2023personalize} and Matcher \cite{liu2023matcher}, and (b) \algo, which is trained on a small data subset for fast convergence and task-specific prompt generation. We report our results on the original test sets for all datasets.

\noindent\textbf{Comparison of \algo with top SAM Prompting Methods.}  PerSAM \cite{zhang2023personalize} and Matcher \cite{liu2023matcher} are the leading SAM-based spatial prompt engineering methods and are a natural comparison to \algo.  \Cref{tab:prompt_engieering} compares performance on nine benchmark datasets. On average, \algo outperforms Matcher by 1.5–2\% on standard computer vision tasks like few-shot segmentation and VOS. The improvement is more substantial on domain-specific tasks, with gains of +35.3\%, +18.5\%, and +5.8\% on TSMU, NWPU VHR-10 and Kvasir-SEG datasets. These results suggest that the prior methods
can struggle in complex scenes with multiple objects, low saliency, varying scales, occlusions, and domain-specific textures. 

\noindent\textbf{Comparison of \algo with Specialist and In-Context Generalist Methods.} We next compare against other generalist (prompt-based) and specialist models (traditional non-promptable models that predict segmentation masks). The results are shown in  \Cref{tab:aerial_seg} and \Cref{tab:few-shot}. On aerial segmentation (\Cref{tab:aerial_seg}), \algo is the only generalist model that can outperform specialist models. It achieves state-of-the-art performance, getting gains of +19.1\% and +10.8\% on TSMU landslide \cite{nagendra2022constructing} and NWPU VHR-10 \cite{su2019object} datasets. We observe from \Cref{tab:few-shot} that \algo significantly outperforms specialist models for few-shot segmentation on COCO-$20^i$ and Pascal-$5^i$ datasets with 53.1\% and 80.4\% mIoU. 
We include comparison with SegGPT but note that its training data includes the test sets of COCO and PASCAL. 
On FSS-1000, \algo achieves comparable performance (89\% mIoU)  to the top specialist model VAT (90.3\$), while have 1/20th as many parameters on 1/5 as much training data. 

\noindent\textbf{Limitations of \algo. } It is also important to study limitations of a model and to understand when it might underperform. We have found that \algo under-performs in medical segmentation and VOS tasks, as shown in \Cref{tab:limitations}.
In the case of Kvasir-SEG dataset, our examination of \algo's prompts suggests that the prompts are good but that SAM struggles with typical characteristcs of medical images, such as  the low contrast of polyp images. Thus, improvements to SAM would lead to improvements to \algo.
For VOS, \algo is not aware of temporal dependencies across video frames, which are crucial for high-accuracy visual object segmentation.  With a one-shot setup, \algo falls short of matching the performance of specialized VOS models that are designed to take this information into account. These limitations highlight areas for further improvement in both \algo and SAM. 

\begin{table}[]
\centering
\resizebox{\linewidth}{!}{%
\begin{tabular}{lccc}
\hline
\multicolumn{1}{c|}{\multirow{2}{*}{Methods}} &
  \multicolumn{1}{c|}{\multirow{2}{*}{\begin{tabular}[c]{@{}c@{}}\# learnable\\ parameters\end{tabular}}} &
  \multicolumn{1}{c|}{TSMU \cite{nagendra2022constructing}} &
  NWPU VHR-10 \cite{su2019object} \\ \cline{3-4} 
\multicolumn{1}{c|}{}                & \multicolumn{1}{c|}{}     & \multicolumn{2}{c}{mIoU (1-shot)} \\ \hline
\multicolumn{4}{l}{aerial segmentation specialist models}                                            \\ \hline
\multicolumn{1}{l|}{U-Net \cite{ronneberger2015unetconvolutionalnetworksbiomedical}}           & \multicolumn{1}{c|}{30M}  & \multicolumn{1}{c|}{64.8}  & 40.8 \\
\multicolumn{1}{l|}{PSPNet \cite{zhao2017pyramid}}          & \multicolumn{1}{c|}{47M}  & \multicolumn{1}{c|}{66.2}  & 43.9 \\
\multicolumn{1}{l|}{DeepLabV3+ \cite{chen2017rethinking}}      & \multicolumn{1}{c|}{43M}  & \multicolumn{1}{c|}{69.1}  & 46.1 \\
\multicolumn{1}{l|}{TSMU \cite{nagendra2022constructing}}            & \multicolumn{1}{c|}{30M}  & \multicolumn{1}{c|}{68.3}  & \textcolor{gray}{n/a}  \\
\multicolumn{1}{l|}{WeSAM \cite{zhang2024improving}}           & \multicolumn{1}{c|}{86M}  & \multicolumn{1}{c|}{\textcolor{gray}{n/a}}   & 35.5 \\
\multicolumn{1}{l|}{PointSAM \cite{liu2024pointsam}}        & \multicolumn{1}{c|}{86M}  & \multicolumn{1}{c|}{\textcolor{gray}{n/a}}   & 59.3 \\ \hline
\multicolumn{4}{l}{in-context generalist segmentation models}                                        \\ \hline
\multicolumn{1}{l|}{Painter (ViT-L) \cite{wang2023images}} & \multicolumn{1}{c|}{354M} & \multicolumn{1}{c|}{32.8}  & 20.8 \\
\multicolumn{1}{l|}{SegGPT (ViT-L) \cite{wang2023seggpt}}  & \multicolumn{1}{c|}{354M} & \multicolumn{1}{c|}{50.6}  & 44.7 \\ \hline
\hline
\multicolumn{1}{l|}{\algo (ours)} &
  \multicolumn{1}{c|}{2.6M} &
  \multicolumn{1}{c|}{\textbf{88.2}} &
  \textbf{70.1} \\ \hline
\end{tabular}
}
\caption{Comparison of \algo with current state-of-the-art specialist and in-context generalist segmentation models on TSMU Landslide \cite{nagendra2022constructing} and NWPU VHR-10 \cite{su2019object} datasets from the task of aerial segmentation.}
\label{tab:aerial_seg}
\end{table}

\begin{table}[]
\centering
\resizebox{\linewidth}{!}{%
\begin{tabular}{lcccc}
\hline
\multicolumn{1}{c|}{\multirow{2}{*}{Methods}} &
  \multicolumn{1}{c|}{\multirow{2}{*}{\begin{tabular}[c]{@{}c@{}}\# learnable\\ parameters\end{tabular}}} &
  \multicolumn{1}{c|}{COCO-$20^i$ \cite{pezoa2016foundations}} &
  \multicolumn{1}{c|}{Pascal-$5^i$ \cite{shaban2017one}} &
  \multicolumn{1}{l}{FSS-1000 \cite{li2020fss}} \\ \cline{3-5} 
\multicolumn{1}{c|}{} &
  \multicolumn{1}{c|}{} &
  \multicolumn{3}{c}{mIoU (1-shot)} \\ \hline
\multicolumn{5}{l}{few-shot segmentation specialist models} \\ \hline
\multicolumn{1}{l|}{HSNet (frozen ResNet-101) \cite{min2021hypercorrelation}} &
  \multicolumn{1}{c|}{2.6M} &
  \multicolumn{1}{c|}{41.2} &
  \multicolumn{1}{c|}{68.7} &
  86.5 \\
\multicolumn{1}{l|}{VAT (ResNet-101) \cite{hong2022cost}} &
  \multicolumn{1}{c|}{52M} &
  \multicolumn{1}{c|}{41.3} &
  \multicolumn{1}{c|}{72.4} &
  \textbf{90.3} \\
\multicolumn{1}{l|}{FPTrans (Vit-B) \cite{zhang2022feature}} &
  \multicolumn{1}{c|}{101M} &
  \multicolumn{1}{c|}{47.0} &
  \multicolumn{1}{c|}{77.7} &
  \textcolor{gray}{n/a} \\ \hline
\multicolumn{4}{l}{in-context generalist segmentation models} &
   \\ \hline
\multicolumn{1}{l|}{Painter (ViT-L) \cite{wang2023images}} &
  \multicolumn{1}{c|}{354M} &
  \multicolumn{1}{c|}{33.1} &
  \multicolumn{1}{c|}{64.5} &
  61.7 \\
\multicolumn{1}{l|}{SegGPT (ViT-L) \cite{wang2023seggpt}} &
  \multicolumn{1}{c|}{354M} &
  \multicolumn{1}{c|}{\textbf{$56.1^{\textcolor{red}{\dagger}}$}} &
  \multicolumn{1}{c|}{$83.2^{\textcolor{red}{\dagger}}$} &
  85.6 \\ \hline
 \hline
\multicolumn{1}{l|}{\algo (ours)} &
  \multicolumn{1}{c|}{2.6M} &
  \multicolumn{1}{c|}{\textbf{53.1}} &
  \multicolumn{1}{c|}{\textbf{80.4}} &
  89.0 \\ \hline
\end{tabular}
}
\caption{Comparison of \algo with current state-of-the-art few-shot specialist and in-context generalist segmentation models on three benchmark few-shot datasets COCO-$20^i$ \cite{nguyen2019feature}, Pascal-$5^i$ \cite{shaban2017one} and FSS-1000 \cite{li2020fss}. $\textcolor{red}{\dagger}$Note that SegGPT was trained on complete COCO and Pascal datasets, including the test sets.}
\label{tab:few-shot}
\end{table}

\begin{table}[]
\centering
\resizebox{\linewidth}{!}{%
\begin{tabular}{lcccc}
\hline
\multicolumn{1}{c|}{\multirow{3}{*}{Methods}} &
  \multicolumn{1}{c|}{\multirow{3}{*}{\begin{tabular}[c]{@{}c@{}}\# learnable\\ parameters\end{tabular}}} &
  \multicolumn{1}{c|}{medical segmentation} &
  \multicolumn{2}{c}{video object segmentation} \\ \cline{3-5} 
\multicolumn{1}{c|}{}                 & \multicolumn{1}{c|}{}     & \multicolumn{1}{c|}{Kvasir-SEG \cite{jha2020kvasir}}    & \multicolumn{1}{c|}{DAVIS-16 \cite{pezoa2016foundations}} & DAVIS-17 \cite{pont20172017} \\ \cline{3-5} 
\multicolumn{1}{c|}{}                 & \multicolumn{1}{c|}{}     & \multicolumn{1}{c|}{mIoU (1-shot)} & \multicolumn{2}{c}{\textit{J \& F} (1-shot)}      \\ \hline
\multicolumn{5}{l}{medical segmentation specialist models}                                                                                        \\ \hline
\multicolumn{1}{l|}{SS-Former-S \cite{shi2022ssformer}}                           & \multicolumn{1}{c|}{30M}  & \multicolumn{1}{c|}{86.8}          & \multicolumn{1}{c|}{\textcolor{gray}{n/a}}      & \textcolor{gray}{n/a}      \\
\multicolumn{1}{c|}{SS-Former-S + PRN \cite{nagendra2024patchrefinenet}}                     & \multicolumn{1}{c|}{32M}  & \multicolumn{1}{c|}{\textbf{89.1}} & \multicolumn{1}{c|}{\textcolor{gray}{n/a}}      & \textcolor{gray}{n/a}      \\ \hline
\multicolumn{5}{l}{video segmentation specialist models}                                                                                          \\ \hline
\multicolumn{1}{l|}{SWEM (ResNet-50) \cite{lin2022swem}} & \multicolumn{1}{c|}{58M}  & \multicolumn{1}{c|}{\textcolor{gray}{n/a}}           & \multicolumn{1}{c|}{91.3}     & 84.3     \\
\multicolumn{1}{l|}{XMem (ResNet-50) \cite{cheng2022xmem} } &
  \multicolumn{1}{c|}{62M} &
  \multicolumn{1}{c|}{\textcolor{gray}{n/a}} &
  \multicolumn{1}{c|}{\textbf{92.0}} &
  \textbf{87.7} \\ \hline
\multicolumn{5}{l}{in-context generalist segmentation models}                                                                                     \\ \hline
\multicolumn{1}{l|}{Painter (ViT-L) \cite{wang2023images}}  & \multicolumn{1}{c|}{354M} & \multicolumn{1}{c|}{37.2}          & \multicolumn{1}{c|}{70.3}     & 34.6     \\
\multicolumn{1}{l|}{SegGPT (ViT-L) \cite{wang2023seggpt}}   & \multicolumn{1}{c|}{354M} & \multicolumn{1}{c|}{43.7}          & \multicolumn{1}{c|}{83.7}     & 75.6     \\ \hline\hline
\multicolumn{1}{l|}{SAMIC (ours)}     & \multicolumn{1}{c|}{2.6M} & \multicolumn{1}{c|}{69.3}          & \multicolumn{1}{c|}{87.2}     & 80.6     \\ \hline
\end{tabular}
}
\caption{Comparison of \algo with current state-of-the-art few-shot specialist and in-context generalist segmentation models on tasks of medical with Kvasir-SEG \cite{jha2020kvasir} dataset and video object segmentation with DAVIS-16 \cite{pezoa2016foundations} and DAVIS-17 \cite{pont20172017} datasets. This table highlights the limitations of using a prompt engineering method like \algo with a VFM like SAM, highlighting the need for specialist models for such hard domains.}
\label{tab:limitations}
\end{table}

\subsection{Ablation Study}\label{sec:ablation}
We perform ablation experiments for establishing our design choice for \algo. We use a sub-sample of Pascal-$5^i$ as a validation set for ablation study.

\begin{table}[htpb]
\centering
\resizebox{\columnwidth}{!}{%
\begin{tabular}{c|cccc}
\hline
\multirow{2}{*}{\begin{tabular}[c]{@{}c@{}}\# 4D conv layers in\\ Hypercorrelation Construction Block\end{tabular}} & \multicolumn{4}{c}{pascal-$5^i$} \\ \cline{2-5} 
           & mIoU (1-shot) & mIoU (5-shot)   & \# learnable parameters & inference time (ms) \\ \hline
1          & 62.1          & 67.6          & 0.2M                    & 7                   \\ 
2          & 65.9          & 70.2          & 0.8M                    & 16                  \\
\textbf{3} & \textbf{66.1} & \textbf{70.4} & \textbf{2.6M}           & \textbf{25}         \\ 
4          & 65.4          & 70.2          & 7.2M                    & 72                  \\ \hline
\end{tabular}
}
\caption{Effect of depth in the Hypercorrelation Construction Block.}
\label{tab:ablation-depth}

\end{table}

\begin{figure}[htpb]
    \centering
    \includegraphics[width=\linewidth]{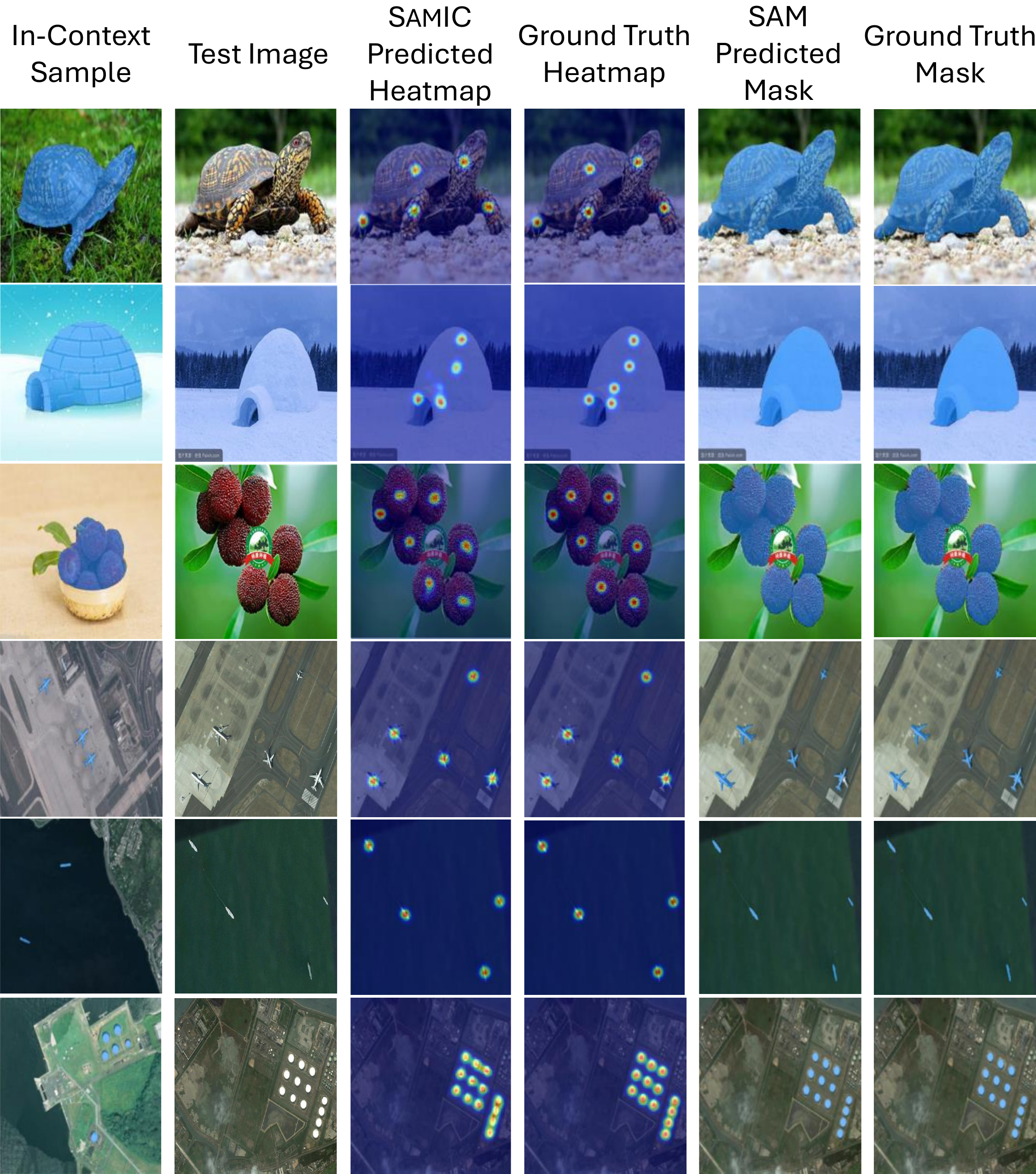}
    \caption{Qualitative results of \algo on FSS-1000 \cite{li2020fss} and NWPU VHR-10 datasets \cite{su2019object}, showing the predicted saliency heatmaps.}
    \label{fig:qual_heatmap}
\end{figure}

\noindent\textbf{Effect of depth in the Hypercorrelation Construction Block.} As shown in \Cref{tab:ablation-depth}, stacking 4D convolutional layers within the hypercorrelation construction block enhances performance up to a depth of three layers, after which performance gains plateau. Increasing depth also raises the number of learnable parameters and inference time. We select an optimal depth of three layers, achieving a model size of 2.6M parameters with a fast inference time of 25 ms, maximizing performance efficiently.

\noindent\textbf{Effect of components of the loss function. } The loss function of \algo comprises three components -- KLD, CC, and NSS, as detailed in \Cref{sec:loss-function}. KLD and CC help align the distributions, while NSS matches the intensity values between predicted and ground truth saliency maps. As shown in \Cref{tab:ablation-loss}, using each component individually yields suboptimal performance. Combining KLD + NSS or CC + NSS improves results, but the highest performance is achieved when all three components are used together.

\begin{table}[htpb]
\centering
\resizebox{0.7\columnwidth}{!}{%
\begin{tabular}{ccc|cc}
\hline
\multicolumn{1}{l}{\multirow{2}{*}{KLD}} &
  \multicolumn{1}{l}{\multirow{2}{*}{CCC}} &
  \multicolumn{1}{l|}{\multirow{2}{*}{NSS}} &
  \multicolumn{2}{c}{pascal-$5^i$} \\ \cline{4-5} 
\multicolumn{1}{l}{} &
  \multicolumn{1}{l}{} &
  \multicolumn{1}{l|}{} &
  \multicolumn{1}{l}{mIoU (1-shot)} &
  \multicolumn{1}{l}{mIoU (5-shot)} \\ \hline
\checkmark &     &     & 58.6          & 61.2          \\
    & \checkmark &     & 52.8          & 53.3          \\
    &     & \checkmark & 42.7          & 43.9          \\
\checkmark &     & \checkmark & 64.9          & 68.1          \\
    & \checkmark & \checkmark & 62.6          & 65.5          \\
\checkmark & \checkmark & \checkmark & \textbf{66.1} & \textbf{70.4} \\ \hline
\end{tabular}
}
\caption{Effect of components of the loss function.}
\label{tab:ablation-loss}

\end{table}

\begin{figure}[htpb]
    \centering
    \includegraphics[width=\linewidth]{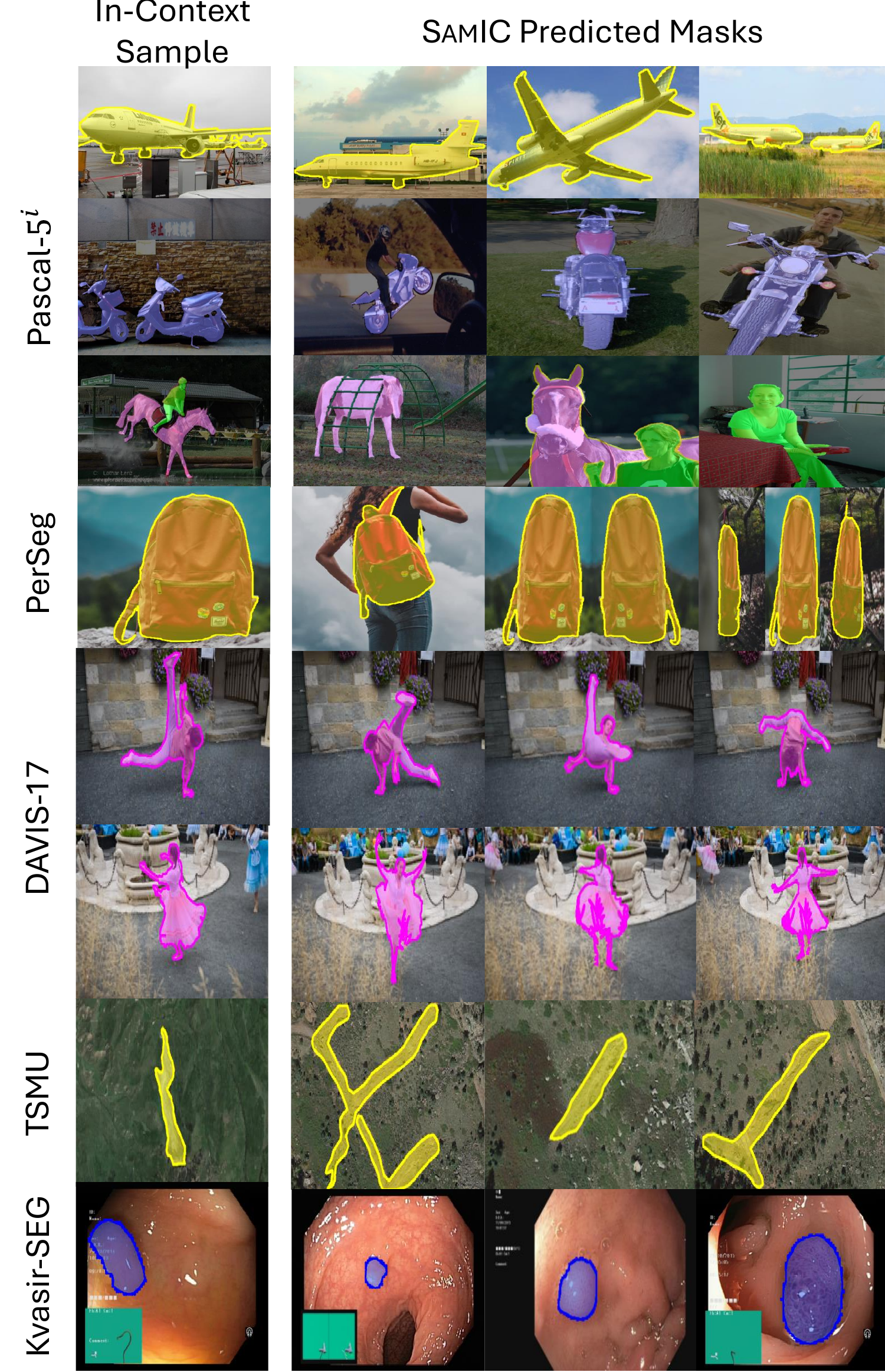}
    \caption{Qualitative results of \algo for 1-shot segmentation on Pascal-$5^i$ \cite{shaban2017one}, PerSeg \cite{ruiz2023dreambooth}, DAVIS-17 \cite{pont20172017}, TSMU \cite{nagendra2022constructing} and Kvasir-SEG \cite{jha2020kvasir} datasets.}
    \label{fig:qual_oneshot}
\end{figure}

\subsection{Qualitative Results}\label{sec:qualitative}
We demonstrate \algo's effectiveness on diverse downstream tasks through qualitative results. In \Cref{fig:qual_heatmap}, the few-shot learner (HSNet) within \algo effectively captures saliency heatmaps, generating high-quality spatial prompts for SAM. \Cref{fig:qual_oneshot} showcases \algo's one-shot performance on the Pascal-$5^i$, FSS-1000, and PerSeg datasets, clearly illustrating \algo’s ability to handle multiple objects (row 3), and perform part-object segmentation based on the given in-context sample (e.g., segmenting only the bike and not the person in row 2). 




\section{Conclusion}\label{sec:conclusion}

This paper introduces \algo, a compact 2.6M parameter network designed to prompt vision foundation models (VFMs) for efficient few-shot segmentation in new domains. Despite its small size, \algo achieves competitive or state-of-the-art performance across few-shot and semantic segmentation benchmarks using minimal training data. By leveraging VFMs, \algo enables any task to be approached as a few-shot learning problem and significantly reduces the cost of creating few-shot segmentation models for new domains. For additional details, please see supplementary material.

{
    \small
    \bibliographystyle{ieeenat_fullname}
    \bibliography{main}
}

\clearpage
\noindent{\Large \textbf{Appendix}}
\appendix

\section{Additional qualitative analysis of embedding similarity approaches}\label{sec:add_qual}
\begin{figure}[htpb]
    \centering
    \includegraphics[width=\linewidth]{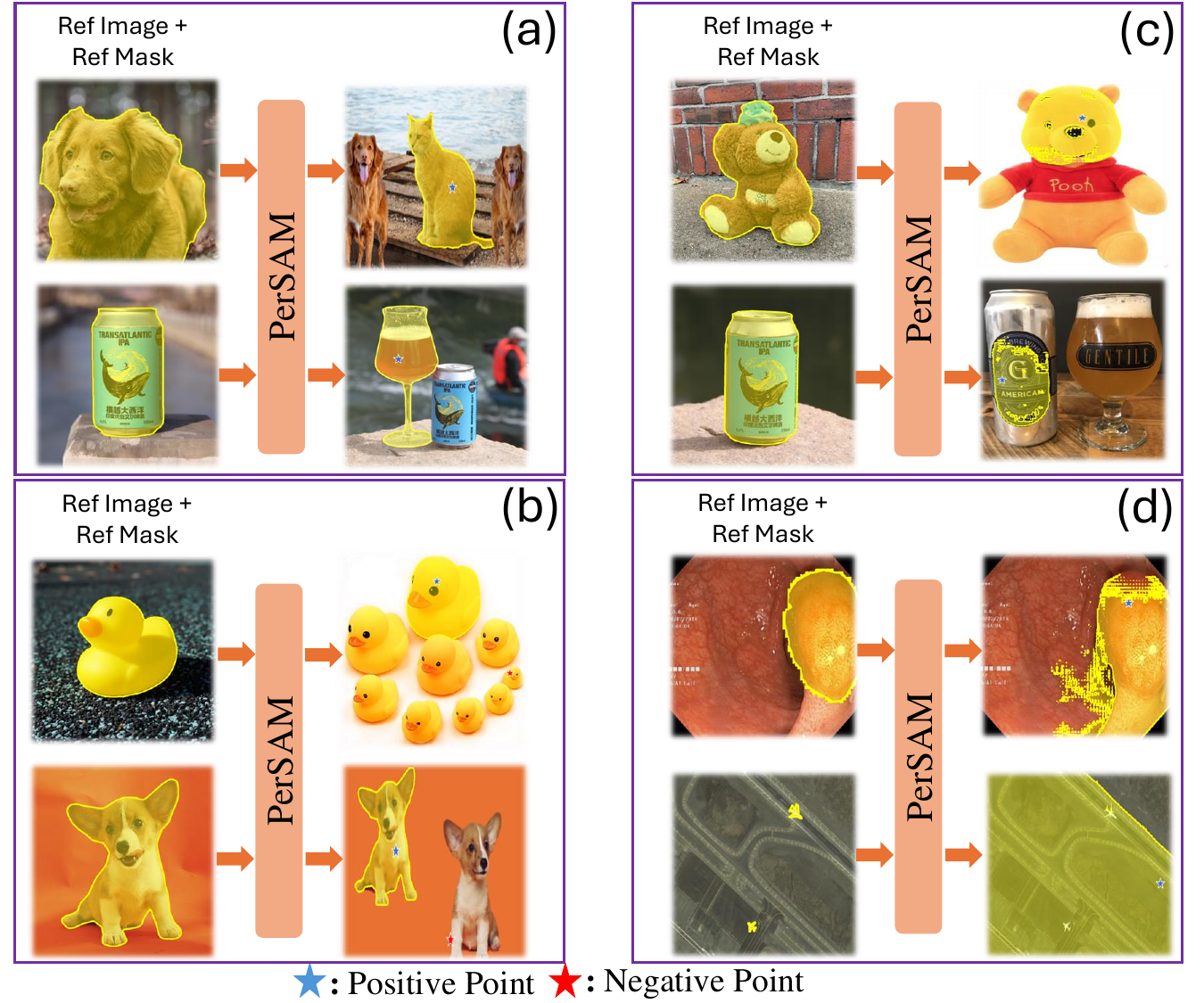}
    \caption{\textbf{Illustration of key limitations of training-free,  embedding similarity-based spatial prompt engineering methods.} 
    }
    \label{fig:persam_problems}
\end{figure}

To further examine the limitations of similarity-based methods, we conduct extensive empirical experiments using PerSAM \cite{zhang2023personalize} as the test bed, as shown in \cref{fig:persam_problems}. PerSAM generates positive (\textcolor{RoyalBlue}{\ding{72}}) and negative (\textcolor{red}{\ding{72}}) point prompts based on an in-context sample consisting of an image and its mask. Our experiments reveal the following key issues: (i) \emph{Susceptibility to visual distractions}—demonstrated in \cref{fig:persam_problems}(a), where correspondence maps may highlight unrelated objects sharing similar features, and positive prompts are often biased toward the image center. (ii) \emph{Poor generalization}—as shown in \cref{fig:persam_problems}(b), where objects with different visual features fail to be detected despite belonging to the same category. (iii) \emph{Inability to handle multiple instances}—highlighted in \cref{fig:persam_problems}(c), where only one instance is typically identified when multiple are present. (iv) \emph{Suboptimal performance on domain-specific tasks}—such as medical or remote sensing segmentation (\cref{fig:persam_problems}(d)). This occurs because (1) generic pre-trained VFMs lack training on domain-specific data, leading to inadequate embeddings, and (2) a single pair of positive-negative prompts results in ambiguous segmentation, as noted in the SAM paper \cite{kirillov2023segment}. 

These findings suggest that relying solely on generic pre-trained VFMs for embedding extraction and using similarity between in-context samples and target images leads to suboptimal prompt generation. Such methods fail to produce task-specific spatial prompts needed for accurately defining object boundaries, differentiating instances, and grouping semantic regions. They also struggle with complex scenes containing multiple objects, varying scales, and occlusions, where generic embeddings lack precision for fine-grained details. Moreover, these approaches often overlook context-specific variations, such as domain-specific textures or unique object features, and cannot adapt effectively to dynamic scenes or novel objects not seen during training. \algo addresses these limitations by learning dense visual correspondences between in-context samples and target images, enabling the generation of more effective, task-specific spatial point prompts tailored to each downstream segmentation task.

\section{Design of \cat}\label{sec:appembedding}

\begin{figure}
    \centering
    \includegraphics[width=\linewidth]{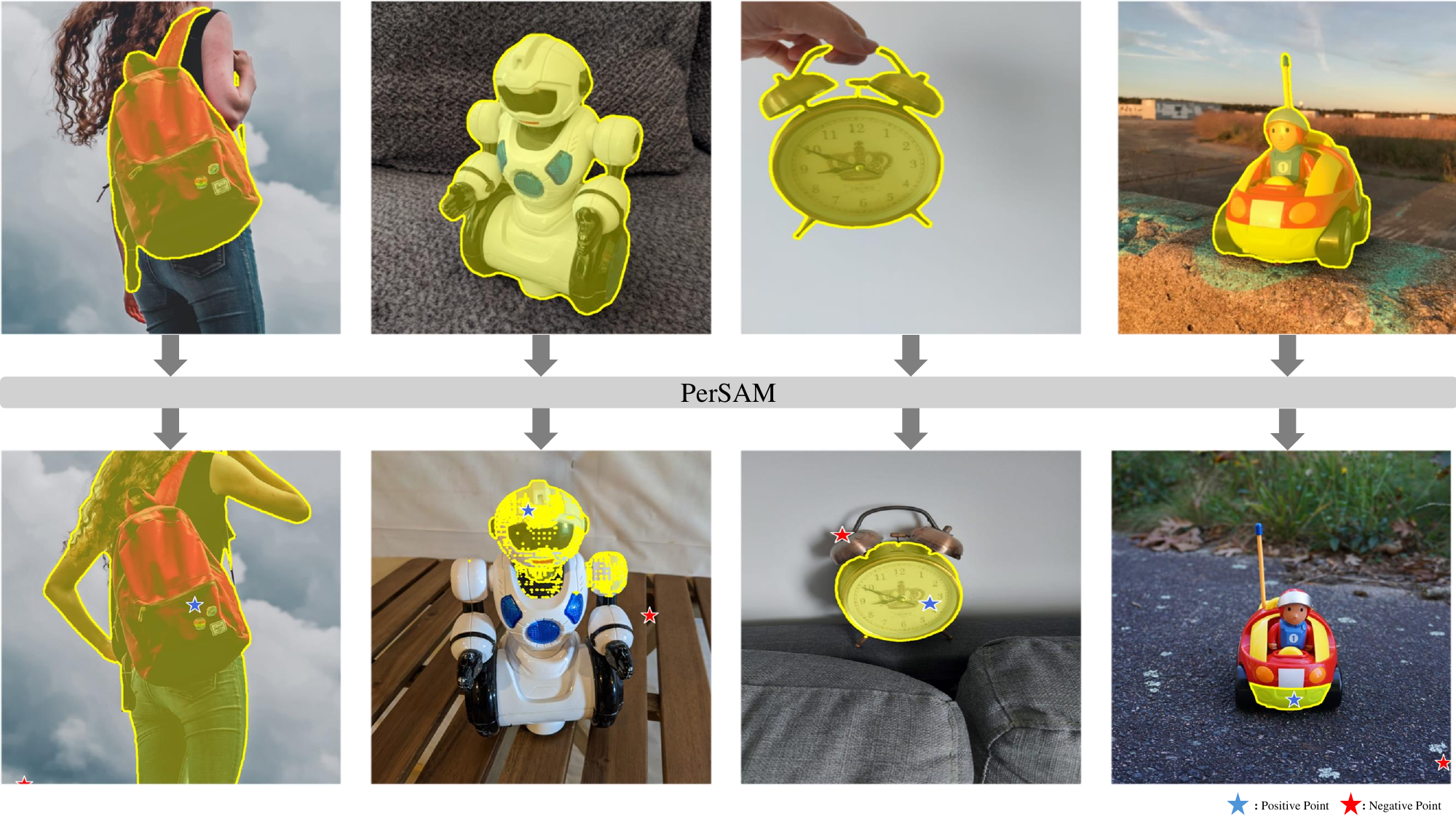}
    \caption{\textbf{Illustration of poor segmentation from insufficient or unclear point prompts.} Results from PerSAM \cite{zhang2023personalize} show that a single point prompt is insufficient for robust segmentation.}
    \label{fig:sam_amb_1}
\end{figure}

\begin{figure*}
    \centering
    \includegraphics[width=\linewidth]{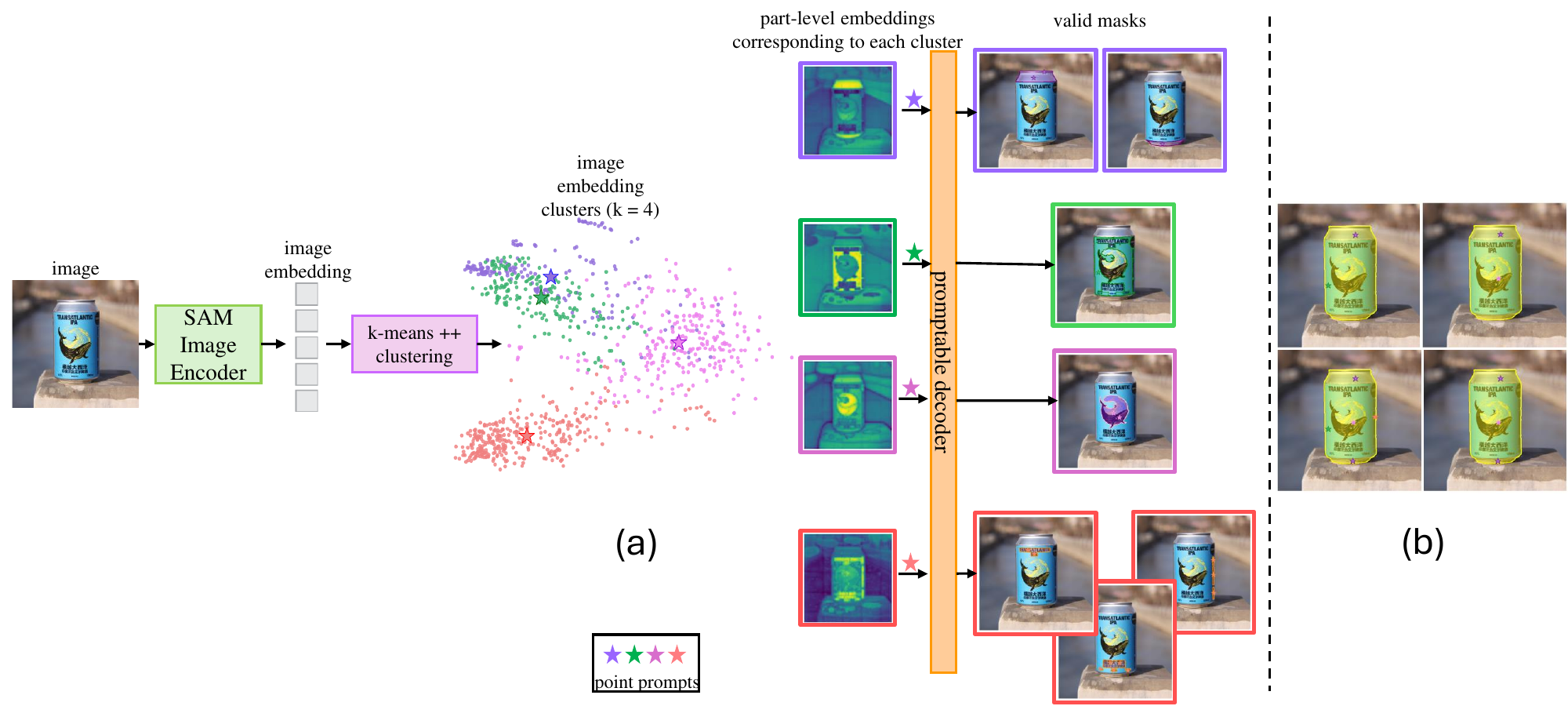}
    \caption{\textbf{(a) Illustration of SAM's ambiguity-aware design.} When the image embedding from SAM is clustered, we get part-level embeddings which represents different textures, colors and text features. A point prompt sampled from one of the clusters would segment only part of the object, leading to ambiguity. This is a byproduct of SAM's design. \textbf{(b) Combination of spatial prompt points corresponding to two or more part-level embeddings.} This figure demonstrates why multiple point prompts are necessary for robust segmentation of objects of interest.}
    \label{fig:sam_ambiguity}
\end{figure*}

SAM’s ambiguity problem arises when provided prompts are unclear or insufficient, leading to misinterpretation, especially in scenarios with multiple objects, unclear boundaries, or occlusions. As illustrated in \cref{fig:persam_problems}(c) and \cref{fig:sam_amb_1}, segmentation masks predicted by PerSAM \cite{kirillov2023segment} show that inadequate prompts often cause SAM to segment only parts of objects or include multiple objects.

To ensure robust data collection, understanding SAM's internal mechanisms, particularly how it processes spatial prompts to generate segmentation masks, is crucial. SAM \cite{kirillov2023segment} features an architecture with an image encoder that generates embeddings, a prompt encoder that processes user inputs, and a lightweight mask decoder that integrates both to predict segmentation masks. SAM is designed to generate a valid mask for any prompt, even ambiguous ones. For instance, a point on a shirt might correspond to either the shirt or the person wearing it. In such cases, SAM produces three masks, ranks them based on predicted confidence scores. and selects the mask with the highest score.

As noted by the authors of the SAM paper \cite{kirillov2023segment}, SAM’s ambiguity decreases with multiple prompts. We illustrate how multiple spatial point prompts help resolve SAM's ambiguity problem in \cref{fig:sam_ambiguity}. To begin, we pass an image through SAM’s image encoder to extract its embedding. We then apply the k-means++ algorithm \cite{arthur2006k} to cluster the image embedding into $n$ parts. These clusters, representing part-level embeddings, are color-coded, with their corresponding centroids marked by colored stars overlaid on each identified group. When we align each cluster in the feature space with pixels in the RGB space, the bright yellow regions in each cluster become prominent. We observe that SAM’s encoder tends to group features based on color and texture, (e.g., the text characters or images on a drink can as shown). Sampling a spatial point prompt from any of the highlighted yellow regions (or from its corresponding cluster in the feature space) results in SAM generating a valid mask, typically segmenting some part of the can, as shown in \cref{fig:sam_ambiguity}(a). However, when we combine multiple spatial point prompts from two or more clusters, SAM successfully segments the entire can, as shown in \cref{fig:sam_ambiguity}(b). 

We embed this part-aware information into \cat to generate robust point prompts as ground truth. SAM assigns lower confidence scores to segmentation maps created with a single point prompt, while the confidence score increases with the addition of multiple prompts, until it eventually saturates. When users manually annotate images, they place spatial point prompts one at a time. The tool displays the current confidence score, which updates with each new prompt. Users can stop when the score saturates, indicating that points covering multiple part-level features have been sampled. This process enables \algo to learn task-specific, ambiguity-aware spatial point prompts.

\section{Additional Ablation Study}

\begin{table}[htpb]
\centering
\resizebox{\columnwidth}{!}{%
\begin{tabular}{l|cccccccccc|c}
\hline
\multicolumn{1}{c|}{\multirow{3}{*}{Methods}} &
  \multicolumn{10}{c|}{pascal-$5^i$} &
  \multirow{3}{*}{\begin{tabular}[c]{@{}c@{}}\# learnable \\ parameters\end{tabular}} \\ \cline{2-11}
\multicolumn{1}{c|}{} & \multicolumn{5}{c|}{1-shot}                                    & \multicolumn{5}{c|}{5-shot}               &       \\ \cline{2-11}
\multicolumn{1}{c|}{} &
  $5^0$ &
  $5^1$ &
  $5^2$ &
  $5^3$ &
  \multicolumn{1}{c|}{mean} &
  $5^0$ &
  $5^1$ &
  $5^2$ &
  $5^3$ &
  mean &
   \\ \hline
PANet \cite{wang2019panet}               & 44.0 & 57.5 & 50.8          & 33.5 & \multicolumn{1}{c|}{40.8} & 55.3 & 67.2 & 61.3          & 53.2 & 59.3 & 23.5M \\
PGNet \cite{zhang2019pyramid}                & 56.0 & 66.9 & 50.6          & 32.4 & \multicolumn{1}{c|}{41.1} & 57.7 & 68.7 & 52.9          & 54.6 & 58.5 & 17.2M \\
PPNet \cite{liu2020part}                & 48.6 & 60.6 & 55.7          & 34.7 & \multicolumn{1}{c|}{43.4} & 58.9 & 68.3 & 66.8          & 58.0 & 63.0 & 31.5M \\
PFENet \cite{tian2020prior}                & 61.7 & 69.5 & 55.4          & 41.2 & \multicolumn{1}{c|}{48.1} & 63.1 & 70.7 & 55.8          & 57.9 & 61.9 & 10.8M \\
RePRI \cite{boudiaf2021few}                & 59.8 & 68.3 & \textbf{62.1} & 52.4 & \multicolumn{1}{c|}{58.0} & 64.6 & 71.4 & \textbf{71.1} & 59.3 & 66.6 & -     \\
HSNet \cite{min2021hypercorrelation} &
  \textbf{64.3} &
  \textbf{70.7} &
  60.3 &
  \textbf{54.0} &
  \multicolumn{1}{c|}{\textbf{59.1}} &
  \textbf{70.3} &
  \textbf{73.2} &
  67.4 &
  \textbf{67.1} &
  \textbf{69.5} &
  \textbf{2.6M} \\ \hline
\end{tabular}
}
\caption{Choosing the few-shot learner for \algo.}
\label{tab:ablation-hsnet}
\end{table}

\noindent\textbf{Choosing Few-shot learner. }The design of \algo requires it to interpret diverse visual cues and establish precise correspondences between a limited set of in-context samples and target images. Furthermore, we prioritize computational efficiency and fast convergence for \algo, making a few-shot learning approach essential. To meet these needs, we chose convolutional architectures, which offer greater efficiency compared to larger transformer-based models. As shown in \cref{tab:ablation-hsnet}, we experimented with multiple convolutional few-shot learners and HSNet \cite{min2021hypercorrelation} performed the best among them. 

\section{Additional Qualitative Results}\label{sec:additional_qual}

\begin{figure*}[h]
    \centering
    \includegraphics[width=\linewidth]{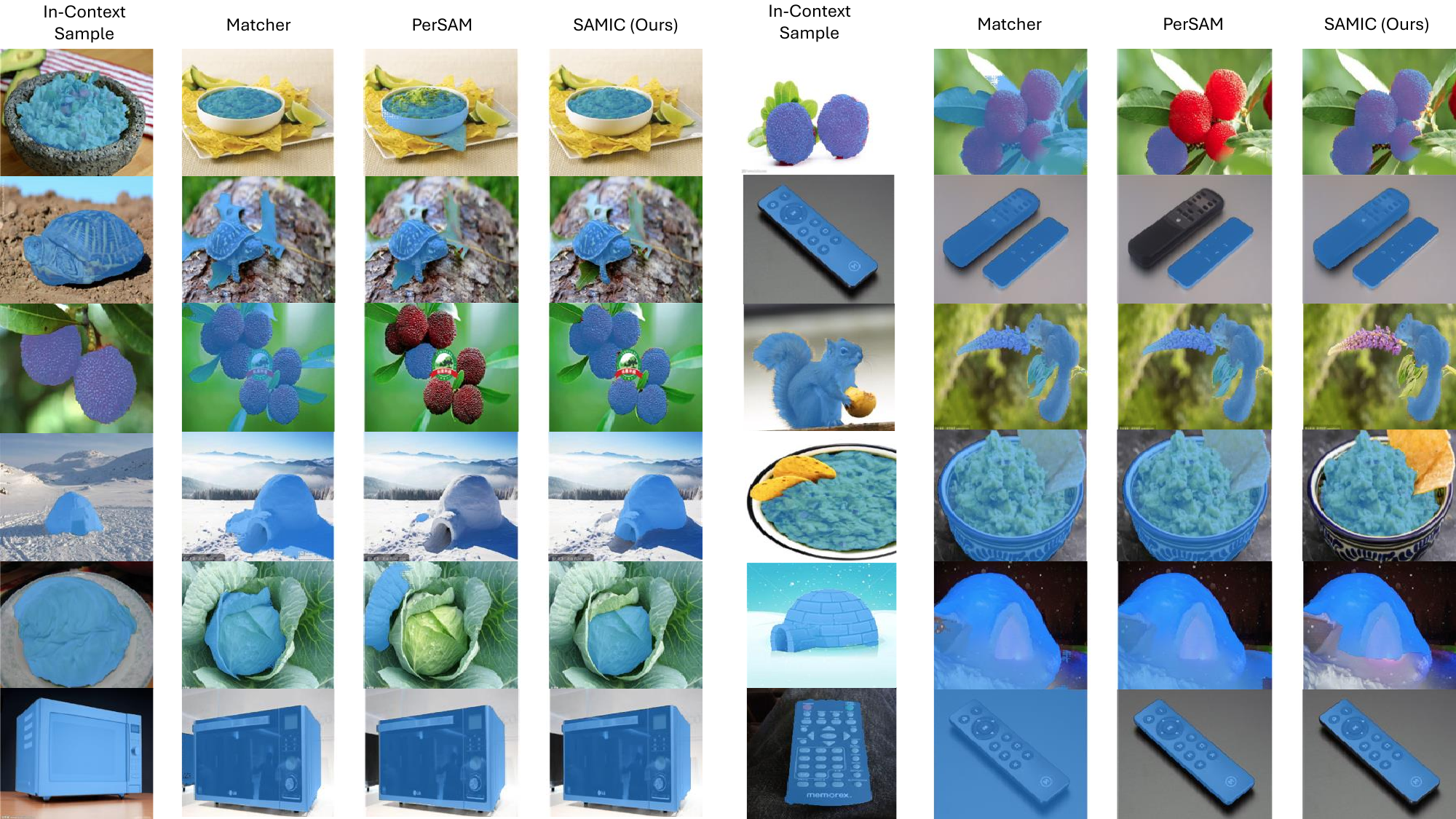}
    \caption{Qualitative results of \algo compared with Matcher and PerSAM on PerSeg dataset.}
    \label{fig:qual_comp1}
\end{figure*}

\begin{figure*}[htpb]
    \centering
    \includegraphics[width=\linewidth]{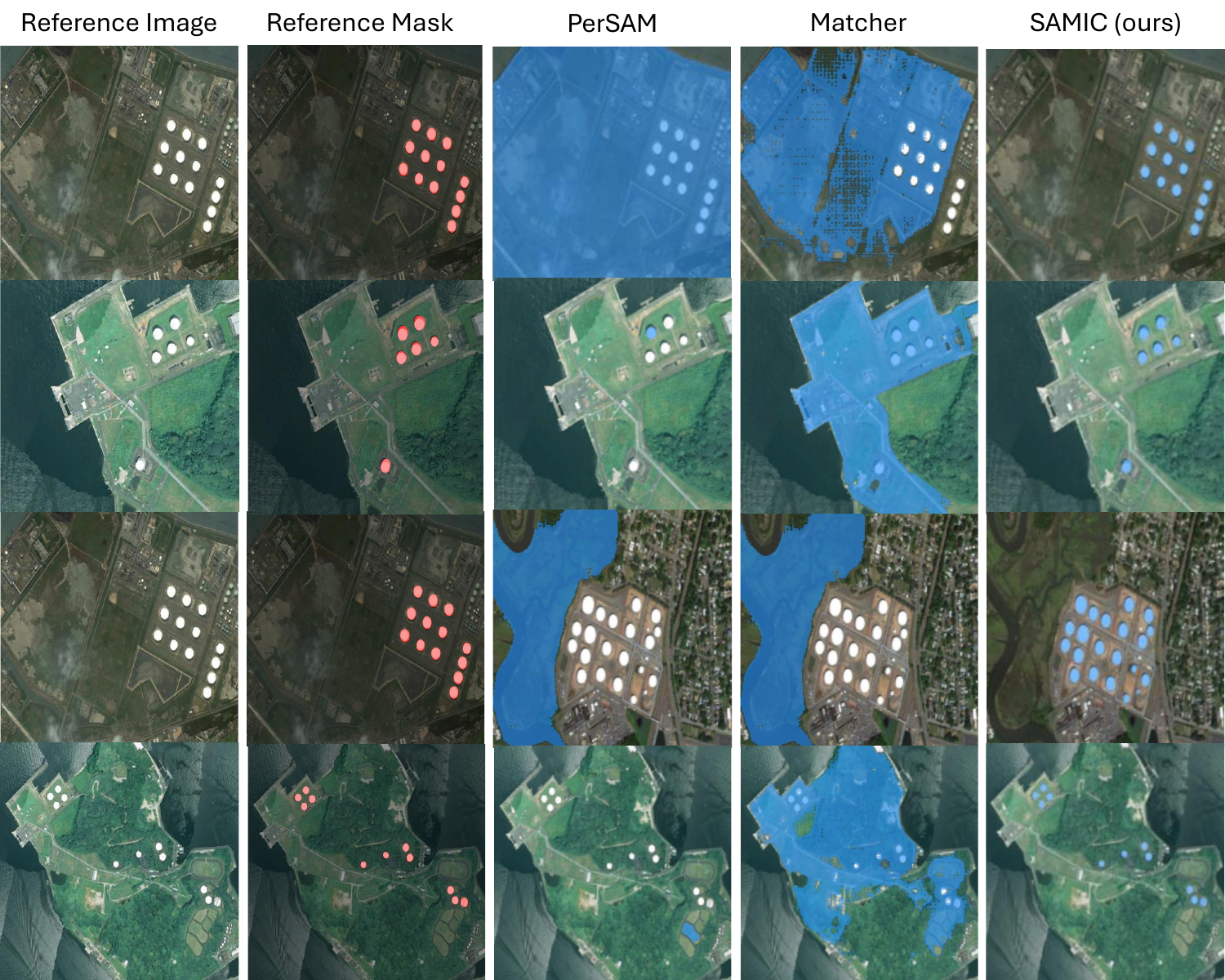}
    \caption{Qualitative results of \algo compared with Matcher and PerSAM on the tanks class from NWPU VHR-10 dataset.}
    \label{fig:qual_comp2}
\end{figure*}

\begin{figure*}[htpb]
    \centering
    \includegraphics[width=\linewidth]{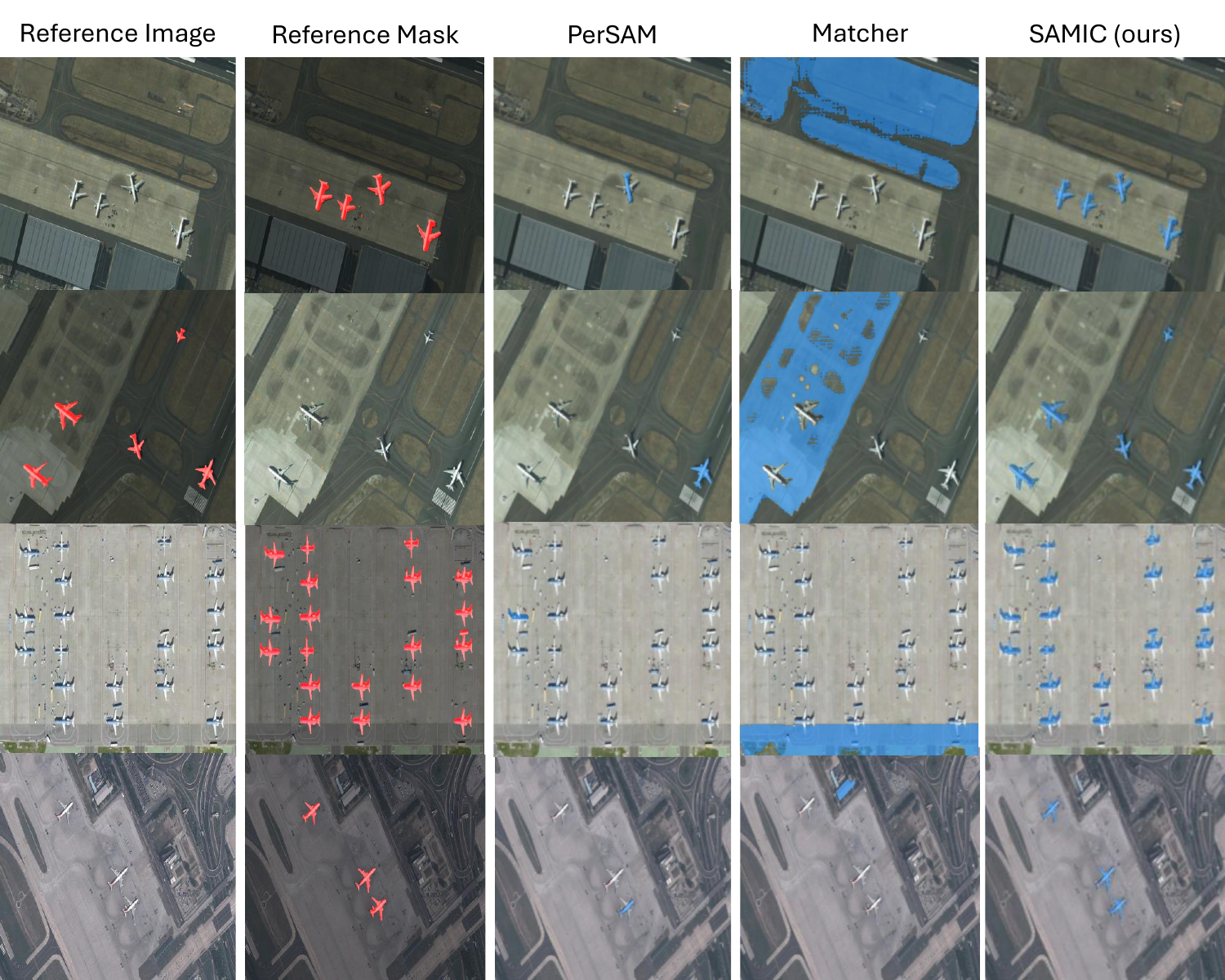}
    \caption{Qualitative results of \algo compared with Matcher and PerSAM on the airplane class from NWPU VHR-10 dataset.}
    \label{fig:qual_comp3}
\end{figure*}

We demonstrate \algo's effectiveness on diverse downstream tasks through additional qualitative results as shown in \cref{fig:qual_comp1}, \cref{fig:qual_comp2} and \cref{fig:qual_comp3}. It can be observed that \algo outperforms contemporary spatial prompt engineering methods PerSAM \cite{zhang2023personalize} and Matcher \cite{liu2023matcher} on PerSeg \cite{ruiz2023dreambooth} and NWPU VHR-10 \cite{su2019object} datasets.


\end{document}